\documentclass[conference]{IEEEtran}
\IEEEoverridecommandlockouts
\usepackage{cite}
\usepackage{amsmath,amssymb,amsfonts}
\usepackage{graphicx}
\usepackage{textcomp}
\usepackage{xcolor}
\usepackage{wrapfig}
\usepackage{subcaption}
\usepackage{multirow}
\usepackage{float}
\usepackage{booktabs}
\usepackage{wrapfig}
\usepackage{mathtools}
\usepackage{algorithm}
\usepackage{algorithmicx}
\usepackage{algpseudocode}

\newcommand{\ours}{NS }
\newcommand{\framework}{FedNS}
\definecolor{colorred}{RGB}{255,150,150} %
\definecolor{colorgreen}{RGB}{150,255,150}    
\definecolor{colorblack}{RGB}{150,150,255}

\def\BibTeX{{\rm B\kern-.05em{\sc i\kern-.025em b}\kern-.08em
    T\kern-.1667em\lower.7ex\hbox{E}\kern-.125emX}}
\begin{document}

\title{Collaboratively Learning Federated Models from Noisy Decentralized Data}


\author{
    \IEEEauthorblockN{Haoyuan Li\textsuperscript{1}, Mathias Funk\textsuperscript{1}, Nezihe Merve Gürel\textsuperscript{2}, Aaqib Saeed\textsuperscript{1}}
    \IEEEauthorblockA{\textsuperscript{1}Eindhoven University of Technology, Eindhoven, Netherlands}
    \IEEEauthorblockA{\textsuperscript{2}Delft University of Technology, Delft, Netherlands}
    \IEEEauthorblockA{\{h.y.li, m.funk, a.saeed\}@tue.nl, n.m.gurel@tudelft.nl}
}

\maketitle

\begin{abstract}
Federated learning (FL) has emerged as a prominent method for collaboratively training machine learning models using local data from edge devices, all while keeping data decentralized. However, accounting for the quality of data contributed by local clients remains a critical challenge in FL, as local data are often susceptible to corruption by various forms of noise and perturbations, which compromise the aggregation process and lead to a subpar global model. In this work, we focus on addressing the problem of noisy data in the input space, an under-explored area compared to the label noise. We propose a comprehensive assessment of client input in the gradient space, inspired by the distinct disparity observed between the density of gradient norm distributions of models trained on noisy and clean input data. Based on this observation, we introduce a straightforward yet effective approach to identify clients with low-quality data at the initial stage of FL. Furthermore, we propose a noise-aware FL aggregation method, namely \textbf{Fed}erated \textbf{N}oise-\textbf{S}ifting (\framework), which can be used as a plug-in approach in conjunction with widely used FL strategies. Our extensive evaluation on diverse benchmark datasets under different federated settings demonstrates the efficacy of \framework. Our method effortlessly integrates with existing FL strategies, enhancing the global model's performance by up to $13.68\%$ in IID and $15.85\%$ in non-IID settings when learning from noisy decentralized data.
\end{abstract}

\begin{IEEEkeywords}
federated learning, decentralized AI, data quality, data-centric machine learning
\end{IEEEkeywords}

\section{Introduction}
\label{sec:intro}
Recent advances in machine learning (ML) have led to a surge in data generated by edge devices. Nonetheless, the ubiquity and heterogeneity of data across these devices pose a significant challenge to the efficacy of training generalizable models. 
In response to this pressing need, federated learning (FL) is gaining traction as a decentralized paradigm in the field of distributed ML. In FL, the model is learned in a decentralized manner where each client device collaboratively trains a global model without directly sharing the local data with the centralized server~\cite{konevcny2016federated, mcmahan2017communication}. 

Despite this, one of the primary challenges in ML is data quality, which directly impacts the performance and reliability of ML models. The complexity and nature of data in the context of deep models, which require large-scale data, significantly amplify these challenges \cite{gudivada2017data}. The multifaceted nature of data quality can be compromised in both the input and feature space, leading to issues such as data incompleteness, feature corruption, and label inconsistency \cite{budach2022effects}. 
Notably, the issue of maintaining high-quality data becomes particularly challenging in FL due to its decentralized nature. FL allows to mitigate domain-specific concerns such as IP security, by imposing the requirement that the server has no visibility into client data, thereby increasing the difficulty of ensuring data quality. Moreover, FL is generally vulnerable to (often non-malicious) client failures, especially when it comes to the noisy model update, which can be seen as a special case of data perturbation \cite{kairouz2021advances}. These failures typically occur on the client side, where the server has no control over. Specifically, unreliable clients unintentionally distort the generalization of the global model by learning from noisy samples or labels with no control over the data collection phase (e.g., due to no direct interest by the federated client). In practice, the utilization of edge devices for FL is a double-edged sword where a large amount of data can be harvested to learn the model, but such data also often entails significant noise contamination. Particularly for tasks of interest like object detection, data collected from image sensors are susceptible to visual distortions, often attributed to the client's lack of technical expertise or environmental interference \cite{dodge2017study}. 

\begin{figure*}[t]
\centering\includegraphics[width=0.75\textwidth]{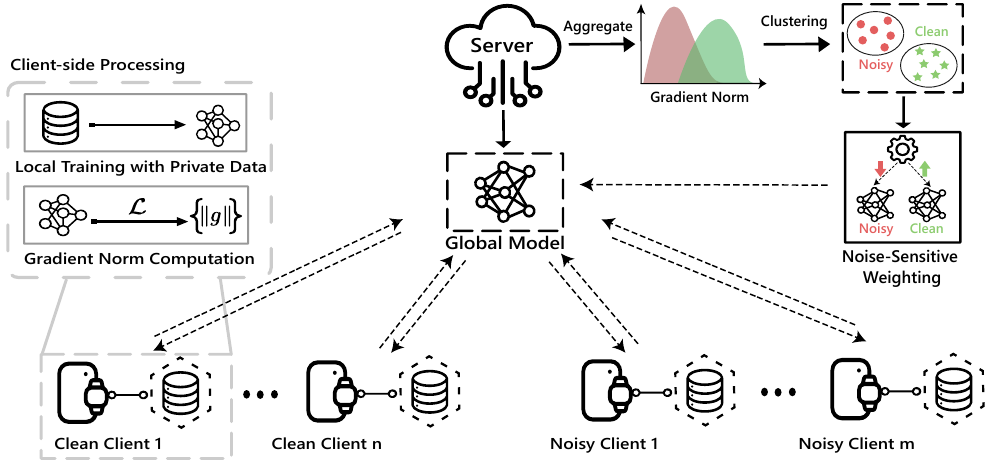}
\caption{\small{Overview of \framework. We propose a plug-in noise-sensitive gradient-norm weighted federated aggregation approach for training high-quality deep models from decentralized data.}} 
\label{fig:overview}
\vspace{-0.5cm}
\end{figure*}
In this work, we mainly focus on the problem of noisy data in the input space, where the features in the client data are (non-maliciously) corrupted. 
To address this, we first investigate the gradient of the loss function, aiming to detect the presence of noisy client data. Our motivation stems from the fact that the gradient space of the model provides an informative signal regarding the usefulness of the data~\cite{gupta2021data} as it directly captures joint information between representations and output~\cite{huang2021importance}. Inspired by this finding, we pose the question \emph{does the gradient provide a meaningful way to distinguish between models of noisy and clean federated clients?} 
From the density of the gradient norm, we see a distinct disparity between models trained on noisy and clean input data. We thus exploit the richness of gradient space to propose a straightforward yet effective approach to identify clients with low-quality data at the beginning stage of FL. We further propose a noise-aware FL aggregation method, namely \textit{Federated Noise Sifting} (\framework). 
Specifically, it re-weights the contributions of client models based on the data quality (i.e., clean or noisy input) while aggregating a global model (see Figure~\ref{fig:overview}). 
\framework can be easily used as a plug-in approach in conjunction with widely used FL strategies. 
To demonstrate the efficacy of \framework, we perform extensive experiments on diverse benchmark datasets under different heterogeneous settings. 
Our findings show that our proposed method effortlessly integrates and works well with existing FL aggregation strategies, such as FedAvg \cite{mcmahan2017communication}, FedProx \cite{li2020federated}, FedTrimmedAvg \cite{yin2018byzantine}, and FedNova \cite{wang2020tackling} which makes it widely applicable.

\textbf{Main contributions}: Our work makes the following key contributions: 
\begin{itemize}
    \item \textbf{\textcolor{black}{Novel single-interaction approach for identifying noisy clients}}. We propose an approach that employs gradient norm analysis to identify and categorize noisy (low-quality model updates) clients at the initial training stage (i.e., a first-round) in a federated context without revealing individual client data for inspection.
    \item \textbf{Robust and plug-in weights aggregation strategy}. We introduce a robust and plug-in weights aggregation strategy to mitigate the adverse impacts of noisy clients, named \framework. It offers a versatile and easily integrable solution to enhance the global model's generalization. We further substantiate the effectiveness of \framework\ through its application to corrupted client data across six datasets and four FL strategies.
    \item \textbf{Systematic conceptualization of noisy FL and benchmark}. We present a concept of noisy FL to understand non-malicious client-side data corruption. Based on this concept, we build\footnote{we will release the code upon acceptance.} the noisy datasets benchmark specifically tailored for FL environments with noisy input.
\end{itemize}

\section{Related work}
\label{sec:related_work}
Dealing with data quality issues is a well-studied problem in machine learning (ML). Significant efforts have been devoted to mitigating the adverse effect of low-quality data on the generalization capabilities of ML models. \cite{dodge2016understanding}, \cite{dodge2017study}, and \cite{NEURIPS2018_0937fb58} reveal that while Deep Neural Networks (DNNs) exhibit advanced generalization abilities but are notably susceptible to low-quality data, particularly to distortions like noise and blur considerably diminishing their performance. To tackle this challenge, previous work enhances DNN robustness to image distortions by selectively correcting the activations of the most noise-susceptible convolutional filters \cite{borkar2019deepcorrect}. Likewise, \cite{liang2020deep} treat image distortion identification as multi-label learning and train a multi-task DNN model to improve the model generalization. Instead of reformulating the architecture of the DNN to improve the robustness of the model, \cite{zheng2016improving} proposes to stabilize the training progress against the input perturbations. In a complementary study, \cite{zhou2017classification} shows that fine-tuning and re-training the model using noisy data can effectively alleviate the effect of image distortion. In addition to this, numerous studies have concentrated on addressing label noise, a form of low-quality data in the label space. Training DNNs on noisy labels tends to overfit the noise patterns, which causes degradation in generalization performance on unseen data.  Most of the existing methods solve the representation learning with noisy labels by estimating the noise transition matrix and innovatively combining methods like adaptation layers, loss corrections, and regularization techniques \cite{patrini2017making, li2019learning, ghosh2017robust, han2020survey}. While these approaches yield substantial results in centralized learning, their applicability is notably limited in federated learning (FL) scenarios, where data is distributed across myriad devices, and the data quality from each client can not be guaranteed due to its private nature.

Recent works in FL mainly focus on addressing the data quality issues pertaining to the label space. Many strategies have been developed to deal with the disparity of label quality in FL. These methods mitigate the impact of label noise by conducting either client selection to re-weight model updates \cite{chen2020dealing, yang2021client, fang2022robust} or data sampling for label correction or exclusion \cite{xu2022fedcorr, yang2022robust, zhang2023noise, zeng2022clc, tsouvalas2024labeling}. Additionally, other works tackle this challenge by correcting the label error. \cite{zeng2022clc} perform label noise correction using consensus-derived class-wise information for dynamic noise identification and label correction. \cite{yang2022robust} tackles label noise in federated learning by updating local models with globally aligned centroids and correcting labels through global model insights. In spite of the progress that has been made in resolving label noise, data subset selection~\cite{gupta2021data}, data valuation~\cite{wang2020principled, li2023data}, dealing with low-quality data in input space (i.e., when noise is in the samples like in images) in the federated setting still remains unexplored. Furthermore, we recognize that client data may be susceptible to backdoor attacks via adversarial methods during both inference and training time \cite{bagdasaryan2020backdoor, wang2020attack}; these concerns fall outside the scope of our investigation. To this end, our method aims for a flexible and efficient solution suitable for a range of FL strategies to learn robust models using decentralized data.

\begin{figure}[t]
    \centering
    \includegraphics[width=0.5\textwidth]{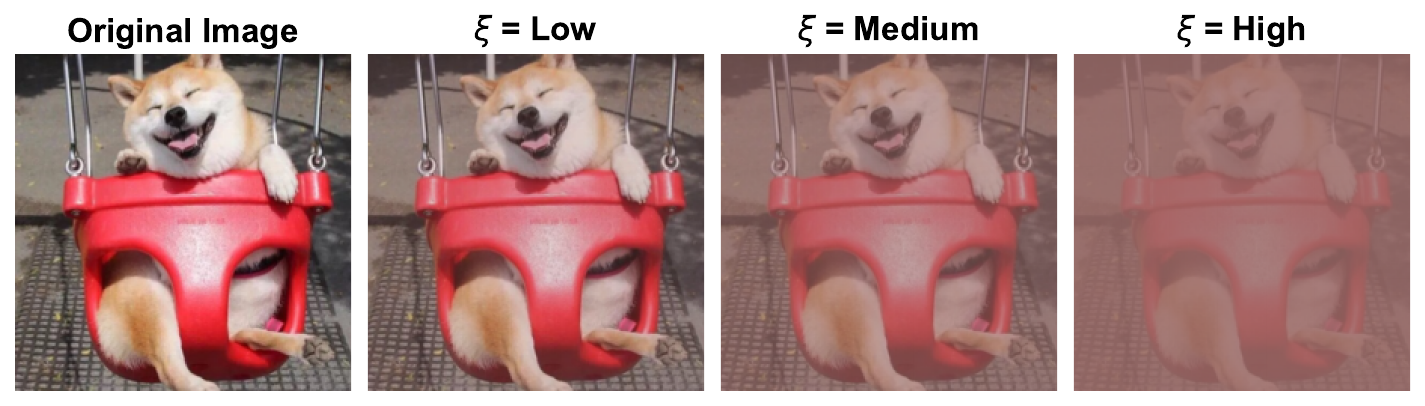}
    \caption{Impact of \textit{contrast} noise on an image with different levels of noise severity.}
    \label{fig:noise_severity}
\vspace{-5mm}
\end{figure}

\section{Method}
We introduce our robust federated learning (FL) method, named \textit{Federated Noise Sifting} (\framework), designed for learning from noisy input. We begin with defining the challenge of FL in the context of input noise and then describe our single-interaction approach to discover and classify noisy and clean clients. Then, we propose a generic noise-aware aggregation method that can be paired with any federated strategy to mitigate the impact of contributions from noisy clients.

\subsection{Problem Definition}
\label{sec:definition}
In a typical FL scenario, a total of $K$ clients collaboratively train a machine learning model by optimizing a global objective function \cite{konevcny2016federated}.
To formulate this setting, consider $K$ disjoint local datasets by clients, where each client indexed by $k$ holds a private dataset $D_k = \{(x_i, y_i)\}_{i=1}^{|D_k|}$, with each pair \((x_i, y_i)\) representing a data sample such that $x_i \in \mathcal{X}$ as the input and $y_i \in \mathcal{Y}$ as the corresponding label. 
 The objective of the clients is to minimize the loss $\mathcal{L}(\theta)$ aggregated from the individual loss of each local client $\ell(\theta, D_k)$ with respect to the aggregated server model parameters $\theta$:
\begin{equation}
\resizebox{\columnwidth}{!}{$
\begin{aligned}
\min_{\theta}\mathcal{L}(\theta) &= \min_{\theta} \frac{1}{\sum_{k \in [K]} {|D_k|}} \sum_{k \in [K]} \sum_{(x_i, y_i) \in D_k} \ell(\theta; \{(x_i, y_i)\}) \\
&= \min_{\theta} \sum_{k=1}^{K} w_k \ell(\theta; D_k)
\end{aligned}
$}
\label{eq:normal_global}
\end{equation}
where 
${w}_k$ denotes the weight assigned to the $k$-{th} client's loss function. Here, we take ${w}_k$ proportional to the size of local data $D_k$ of client $k$ such that ${w}_k = |D_k| / \sum_{k=1}^{K} |D_k|$, similar to that of \cite{mcmahan2017communication}.

\textbf{Noisy Federated Learning} \ \  We focus on a noisy federated learning setting where perturbations are present in the input space $\mathcal{X}$, that often occur due to the unexpected variability in data collection or physical conditions and other environmental factors \cite{singh2016comparative, tian2020deep, villar2021deep}. 

Consider an FL setup consisting of $K$ participating clients, where $M$ out of $K$ clients' local datasets are partially contaminated by some noise, whose function is defined as $\eta(\cdot): \mathcal{X}\mapsto \mathcal{X}$. 
Conversely, the local datasets of the remaining $N$ clients, where $N=K-M$, are unaffected.
To simplify the exposure, we index the noisy clients by $m$ and clean clients by $n$, and denote the respective client data by $D_m$ and $D_n$, where $\{\cup_m D_m\} \cup \{\cup_n D_n\}= \cup_k D_k$. 
We assume that the local dataset of each noisy client $m$ can be further split disjointly into clean data $D_{m}^{\text{clean}}$ and noisy data ${D}_{m}^{\text{noisy}}$ such that $D_{m}^{\text{clean}}\cup D_{m}^{\text{clean}} = D_m$ and $D_{m}^{\text{clean}}\cap D_{m}^{\text{clean}} = \emptyset$.  
In such noisy federated learning setting, the global objective can be expressed as:
\begin{equation}
\begin{split}
\min_{\theta}\mathcal{L}(\theta) &= \min_{\theta} \hspace{-.5em} \sum_{\substack{n \in [K]\\ \textrm{s.t. } D_n \textrm{ is clean}}} \hspace{-1.5em}{w}_n\ell(\theta; D_n)\hspace{0.5em}+\hspace{-1em}\sum_{\substack{m \in [K]\\ \textrm{s.t. } D_m \textrm{ is noisy}}} \hspace{-1.5em}{w}_m{\ell}(\theta; D_m)\\
&= \min_{\theta} \hspace{-.5em} \sum_{\substack{n \in [K]\\ \textrm{s.t. } D_n \textrm{ is clean}}} \hspace{-1.5em}{w}_n\ell(\theta; D_n)\\
&\quad+\hspace{-1em}\sum_{\substack{m \in [K]\\ \textrm{s.t. } D_m \textrm{ is noisy}}} \hspace{-1em} {w}_m \left( \frac{|D_{m}^{\textrm{clean}}|}{|D_m|}{\ell}(\theta; D_{m}^{\textrm{clean}})\right.\\
&\quad\left.+\frac{|D_{m}^{\textrm{noisy}}|}{|D_m|}{\ell}(\theta; D_{m}^{\textrm{noisy}}) \right).
\end{split}
\label{eq:noisy_global}
\end{equation}

\begin{figure}[h]
\vspace{-0.4cm}
\centering
\includegraphics[width=0.35\textwidth]{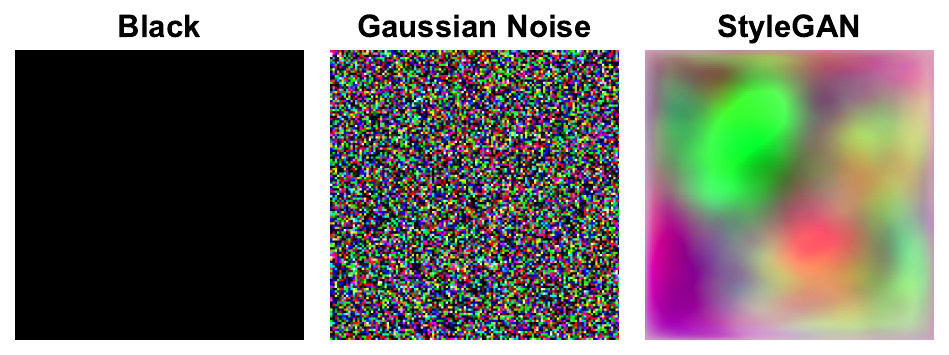}
\caption{Illustration of patch-based data corruption.}
\label{fig:patch_data}
\vspace{-0.2cm}
\end{figure}

\textbf{Noisy Input Data} \ \ In decentralized data collection, input corruption typically occurs on the client-side, influenced by varying degrees of real-world noise and varying degrees of corruption from weak to strong. To address this, we propose a definition for noisy FL data that accounts for the intensity and level of noise in each client's local datasets.


Consider the partition of a global training dataset $D$ into $K$ local datasets $D_k$ for $k\in[K]$. This partitioning follows either an Independent and Identically Distributed (IID) or a non-IID setting~\cite{mcmahan2017communication}. Within this set of $K$ clients, we randomly pick up $M$ clients as the noisy clients, in which the local datasets of these clients are contaminated by input noise, such as Gaussian blur, contrast, defocus blur (see the extended list in the Section \ref{sec:exp_data}). Consequently, the local dataset of $m$-{th} noisy client $D_m$ is transformed into noisy local data by applying a randomly selected transformation $\tau$ from all the data transformation $\mathbf{T}$ with a specific severity level $\xi \in \{low, medium, high\}$. This can be expressed as:
\begin{equation}
    {D}_m^{\textrm{noisy}} = \left\{\ \eta(x, \tau, \xi) \mid x \in D_m, \tau \in \mathbf{T} \right\}.
\end{equation}
Thus, we define the noise level of client $m$ as $\emph{NL}_m = \frac{|{D}_m^{\textrm{noisy}}|}{|D_m|}$, which represents the fraction of noisy corrupted data samples over the entire local data of client $m$. We further introduce image noise severity to quantify the extent of distortion in digital images. Severity levels are scaled from low to high, indicating the intensity of the noise. An illustration of different levels of image noise severity is shown in Figure \ref{fig:noise_severity}.

In addition to the image distortion, we further introduce patch-based image corruption to simulate the condition when the nullified data is injected into the client data. We term this injection of nullified images as \textit{patch-based corruption}. We characterize these corruptions in the following three forms (but note that many others could exist): black patches, patches with Gaussian noise, and generative patches (we use StyleGANv2) \cite{baradad2021learning}, illustrated in Figure \ref{fig:patch_data}. Here, specific data samples in the client's local dataset are substituted with such patches, while the corresponding labels remain unchanged, e.g., a random generative patch has an object label assigned to it.

\begin{figure}
 \centering
 \begin{subfigure}[b]{0.24\textwidth}
     \centering
     \includegraphics[width=\textwidth]{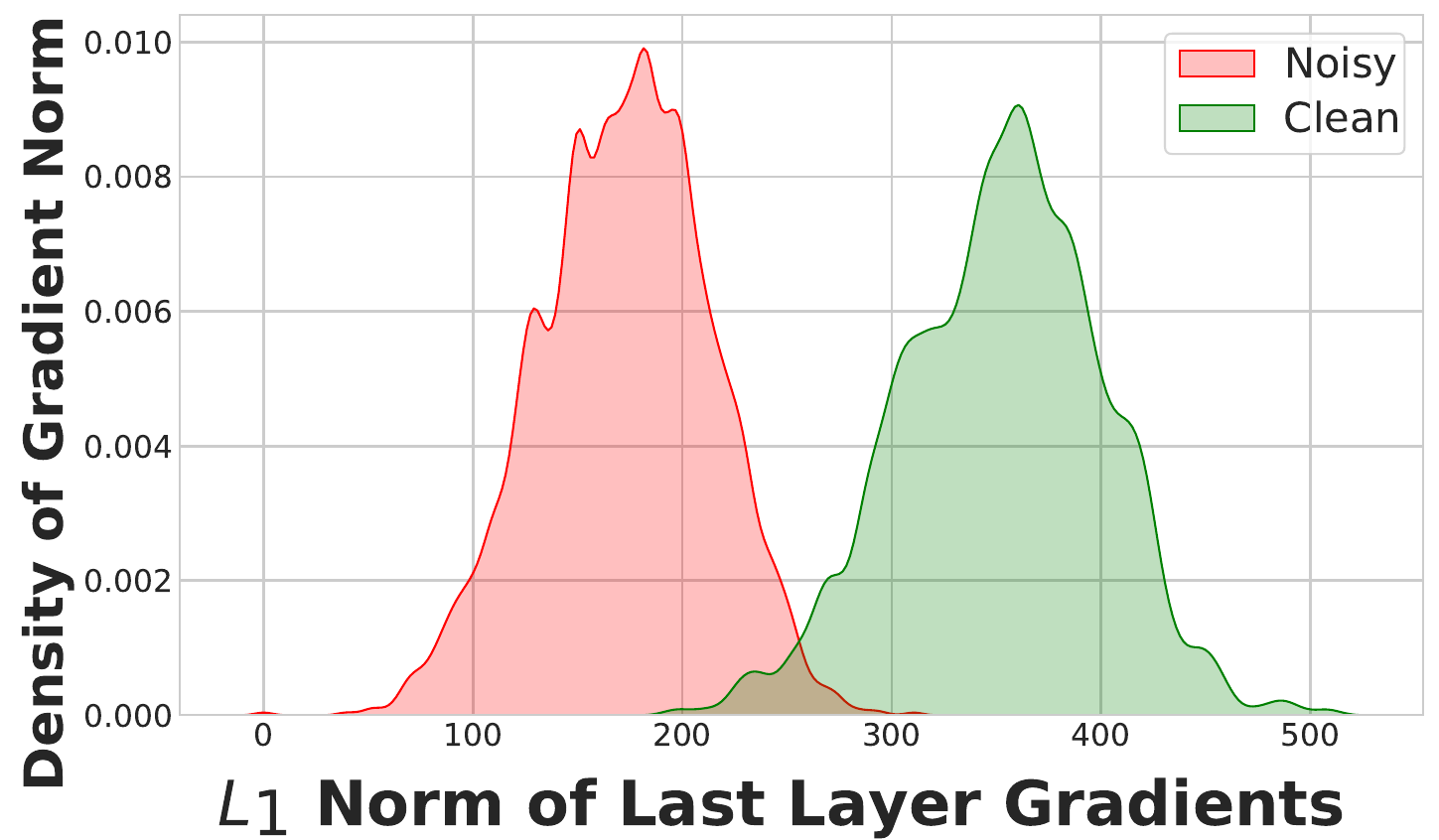}
     \caption{CIFAR-10}
     \label{subfig:kde-cifar10}
 \end{subfigure}
 \hfill
 \begin{subfigure}[b]{0.24\textwidth}
     \centering
     \includegraphics[width=\textwidth]{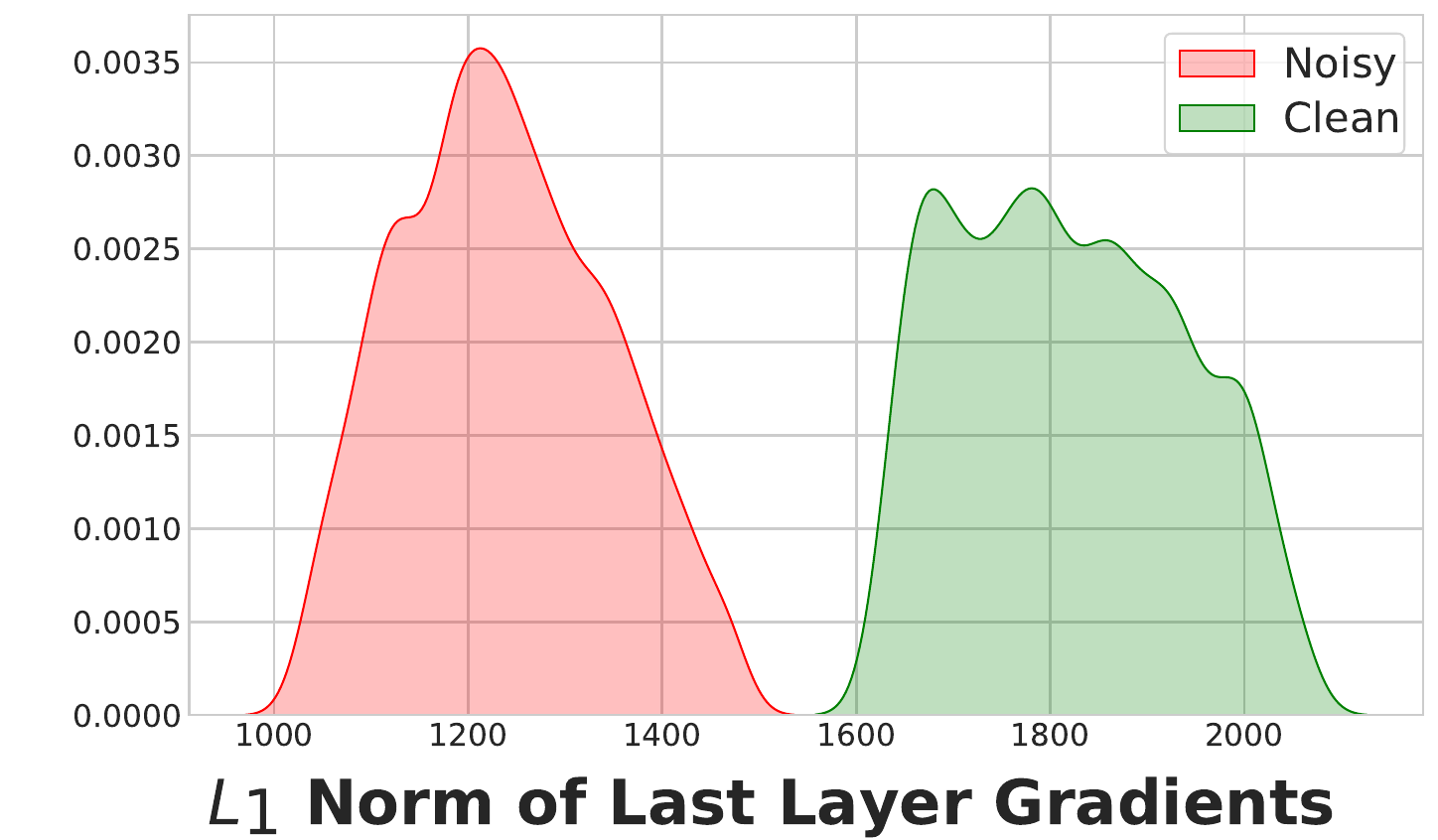}
     \caption{PathMNIST}
     \label{sub:kde-pathmnist}
 \end{subfigure}
    \caption{\small{Gradient norm density during federated training on Fashion-MNIST (15 clean, 5 noisy clients, $\emph{NL}$=100\%) using mean aggregation.}}
    \label{fig:kde}
    \vspace{-5mm}
\end{figure}

\subsection{Single-Interaction Discovery of Noisy Clients}
Unlike centralized learning,  where the server has direct access to inspect all the data, FL presents a unique challenge due to the decentralized and private nature of the data. In FL, clients contribute to the server by sharing their model parameters rather than raw data \cite{mcmahan2017communication, konevcny2016federated}, making it impossible to visually verify the quality of data. To address this issue, we propose a novel approach that can discover the noisy client with only a single-interaction (or in a one-shot manner) using only a local model. Specifically, we identify the noisy clients by evaluating the gradient norms~\cite{paul2021deep,huang2021importance} calculated from local data during the initial training round without compromising the client's data confidentiality.

\begin{figure}[t]
 \centering
 \begin{subfigure}[b]{0.24\textwidth}
     \centering
     \includegraphics[width=\textwidth]{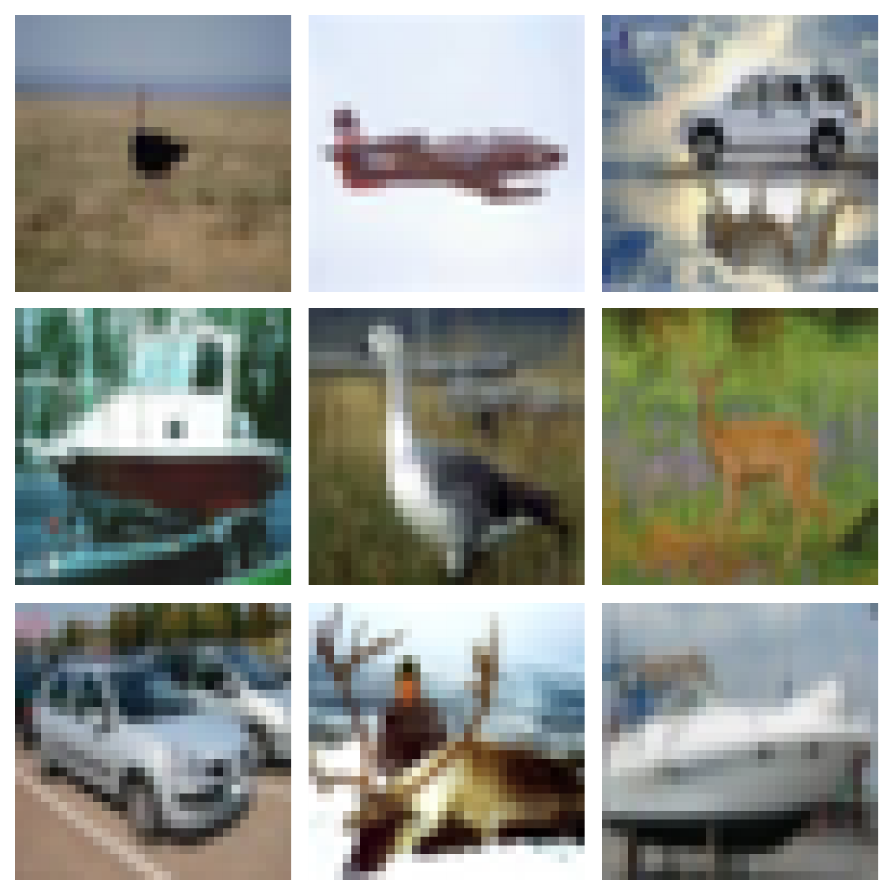}
     \caption{Samples from clean clients}
     \label{subfig:data_sample_clean}
 \end{subfigure}
 \hfill
 \begin{subfigure}[b]{0.24\textwidth}
     \centering
     \includegraphics[width=\textwidth]{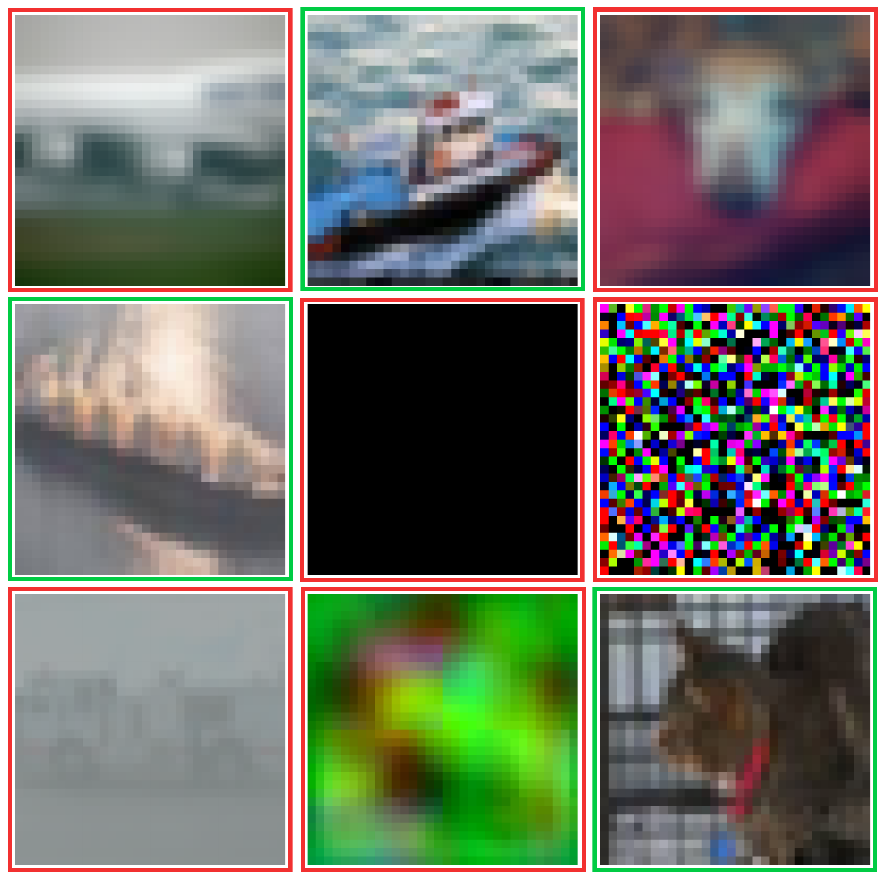}
     \caption{Samples from noisy clients}
     \label{subfig:data_sample_noisy}
 \end{subfigure}
    \caption{Qualitative analysis of clean and noisy CIFAR-10 samples identified by~\framework. Nine images from each cluster are shown. In Figure \ref{subfig:data_sample_noisy}, noisy samples exhibit mixed corruptions. \textcolor{red}{Red} and \textcolor{green}{green} grids denote noisy and clean images, respectively.}
    
    \label{fig:qualitative_analysis_data}
    \vspace{-8mm}
\end{figure}

The challenge in training deep neural networks on input (such as images) with noise and distortions is that the model can easily affected by the perturbations, thus producing overconfident predictions on distorted input data samples \cite{dodge2016understanding, dodge2017study, NEURIPS2018_0937fb58, zheng2016improving} or leading to poor generalization. \textcolor{black}{Motivated by the findings of~\cite{gupta2021data}, we empirically discover that a client's local model trained on distorted inputs exhibits distinct behavior in its gradient space compared to a model trained on clean inputs (see Figure~\ref{fig:kde}).}
In other words, the magnitude of gradients, when trained on clean or corrupted data, is indicative of variation in the model's learning process. The gradient norm effectively encapsulates the relationship between input features and the model’s output. Intuitively, we propose a single-interaction method to estimate data distortion by evaluating the gradient norm of the softmax cross-entropy loss $\mathcal{L}$ during the backpropagation phase of local model training that we get at no extra cost. Given the model parameterized by $\theta$, we define the loss of the input batch data $\mathbf{x}$ with its ground true labels $\mathbf{y}$ as:
\begin{equation}
    \mathcal{L}(\theta; \mathbf{x}, \mathbf{y}) = -\log \left( \frac{e^{f_\mathbf{y}(\theta; \mathbf{x}) / T}}{\sum_{c=1}^C e^{f_c(\theta; \mathbf{x}) / T}} \right)
\end{equation}
where the $f(\cdot)$ signifies the model's output function, $f_y(\theta; \mathbf{x})$ is the output logit from the model $f$ for the ground true label $\mathbf{y}$ given the input $\mathbf{x}$, and 
$f_c(\theta; \mathbf{x})$ is the output logit from the neural network for class $c$. The gradient norm of the input batch data $\mathbf{x}$ over $C$ classes is defined as follows:
\begin{equation}
     g(\mathbf{x}) = \left\| \nabla_{\theta} \mathcal{L}(\theta; \mathbf{x}, \mathbf{y}) \right\|_p
\end{equation}
The term $g(\mathbf{x})$ refers to the gradient norm of the input batch data $\mathbf{x}$. Considering the $k$-{th} client within a set of clients, the overall local dataset of client $D_k$ is split into $B$ mini-batches. Hence, we compute the gradient norms for the entire local dataset $D_k$, expressed as $\{ g(\mathbf{x}_b) \}_{b=1}^B$. After the initial round of local training, the server aggregates the gradient norm results from each of the $N$ clients, forming a score vector by computing variance. By applying the kernel density estimation on the aggregated gradient norm vectors from the clients, we observe a distinct separation in the distributions of clean and noisy clients in Figure \ref{fig:kde}. Subsequently, we apply the K-means clustering on the set of all client gradient norm vectors to distinguish two clusters, as illustrated in Figure \ref{fig:overview}. We then compute the centroid of each cluster, categorizing the one with the higher centroid value as `clean' and its counterpart as `noisy' inspired by the findings of~\cite{huang2021importance}. We examine the outcomes of client clustering within the~\framework~framework. During the initial training round, the server categorizes all clients into either clean or noisy clusters based on gradient norms. We then visualize 9 images sampled from both clean and noisy clusters. As shown in Figure \ref{fig:qualitative_analysis_data}, the images from the noisy cluster are contaminated by a mixture of noise, whereas those from the clean cluster remain intact.

A key advantage of our proposed method is its implementation at the beginning of the federated training process, eliminating the need for additional communication rounds. By utilizing only the aggregated gradient norms from each client, our approach is directly usable (as a plug-in) with various FL strategies. Our method incorporates only the gradient norm (i.e., a scalar value) along with standard model parameters during the aggregation phase, thereby not increasing communication overhead. Moreover, this method enhances security by requiring clients to share only scalar values, significantly reducing the risk of sensitive data leakage.

\begin{table*}[t]
\centering
\caption{\small{Comparison of average accuracy across three independent runs for different datasets under clean and noisy client data scenarios. For the noisy data scenario, we consider 5 clean clients and 15 noisy clients with 100\% noise level. Models are trained with FedAvg.}}
\label{tab:NoiseImpact}
\resizebox{\textwidth}{!}{%
\begin{tabular}{@{}lcccccccccccc@{}}
\toprule
\multirow{2}{*}{\textbf{Data}} & \multicolumn{2}{c}{\textbf{CIFAR10}} & \multicolumn{2}{c}{\textbf{CIFAR100}} & \multicolumn{2}{c}{\textbf{PathMNIST}} & \multicolumn{2}{c}{\textbf{FMNIST}} & \multicolumn{2}{c}{\textbf{EuroSAT}} & \multicolumn{2}{c}{\textbf{Tiny-ImageNet}} \\
\cmidrule(r){2-3} \cmidrule(r){4-5} \cmidrule(r){6-7} \cmidrule(r){8-9} \cmidrule(r){10-11} \cmidrule(r){12-13} 
& \textbf{IID} & \textbf{Non-IID} & \textbf{IID} & \textbf{Non-IID} & \textbf{IID} & \textbf{Non-IID} & \textbf{IID} & \textbf{Non-IID} & \textbf{IID} & \textbf{Non-IID} & \textbf{IID} & \textbf{Non-IID} \\
\midrule
Clean & 90.14\% & 85.52\% & 64.79\% & 62.36\% & 87.74\% & 82.55\% & 92.34\% & 89.37\% & 94.72\% & 95.12\% & 53.26\% & 52.88\% \\
Noisy & 78.62\% & 73.51\% & 44.58\% & 42.10\% & 54.80\% & 52.14\% & 88.14\% & 84.67\% & 67.39\% & 75.06\% & 24.32\% & 22.90\% \\
\bottomrule
\end{tabular}
}
\end{table*}

\subsection{Federated Noise Sifting}
In FL, the formulation of the global model normally involves the aggregation of local models from each client, weighted based on the total count of data samples in their respective local datasets \cite{konevcny2016federated, mcmahan2017communication}. When applying an average aggregation strategy when clients' data follows the IID setting, the contribution of each client's model parameters to the server is equivalent. Consequently, in scenarios of noisy federated learning, the contribution of noisy clients can seriously degrade the generalization quality of the server-side aggregated model, as detailed in Table \ref{tab:NoiseImpact}. 

Normally, the optimization function of a standard noisy FL problem is to minimize the loss from both noisy and clean clients defined in Equation \ref{eq:noisy_global}, where the model parameters from each client's model are weighted by the amount of data it has divided by the total number of data samples in all clients. To mitigate the detrimental effect of noisy clients, we propose a noise-sensitive weights aggregation strategy that is resilient to data corruption during the federated training process, termed \texttt{\framework}. Specifically, we introduce the weighting factor $\alpha$ and $\beta$ to control the contribution of model updates from clean and noisy clients, respectively. Given a set of client updates and their corresponding weights, the global parameter $\theta$ is updated by aggregating the weighted updates from all clients, including both clean and noisy ones. Considering $\theta_g^t$ as the global parameter at the training round $t$, $K$ total clients including $N$ clean clients and $M$ noisy clients. Let ${w}_k$ be the weight and $\theta_k$ be the model parameter of client $k$. Thus, We formulate the proposed global parameter update as follows:
\begin{equation}
\begin{aligned}
\theta_g^t \leftarrow \theta_g^{t-1} - \eta \bigg( & \sum_{\substack{n \in [K] \\ \textrm{s.t. } D_n \textrm{ is clean}}} \mathllap{\smash{w}_n} \alpha \nabla \ell(\theta_n^{t}; D_n) \\
&+ \sum_{\substack{m \in [K] \\ \textrm{s.t. } D_m \textrm{ is noisy}}} \mathllap{\smash{w}_m} \beta \nabla {\ell}(\theta_m^{t}; D_m) \bigg)
\end{aligned}
\label{eq:global_update}
\end{equation}
where $\nabla \ell(\theta_n^{t}; D_n)$ represents the gradient of the loss function for the clean client $n$ at training round $t$, while $\nabla {\ell}(\theta_m^{t}; D_m)$ denotes the corresponding gradient for the noisy client $m$ in the same training round. $\eta$ denotes the learning rate. The coefficients $\alpha$ and $\beta$ are utilized to proportionally scale the model update from clean and noisy clients, respectively. Concisely, our approach involves assigning a higher weighting factor to clean clients and a reduced one to noisy clients to lower their contributions to the overall global model. We provide the complete \framework~ algorithm in Algorithm \ref{algorithm}. 

\begin{algorithm*}[ht]
\scriptsize
\caption{\texttt{Federated Noise Sifting (\framework)}: Our method can be easily plugged into various FL strategies like the standard Federated Averaging. \( N \) is local clients, \( T \) is the total training rounds, \( E \) is the epoch of local training, and \( \eta \) is the learning rate. At each round \( t \), the server will sample \( S_t \) clients that can be divided into \( n \) clean clients and \( m \) noisy clients. (\colorbox{colorred}{\textbf{FedAvg}}, \colorbox{colorgreen}{\textbf{\framework}})}
\begin{algorithmic}[1]
\Statex \textbf{Server executes:}
\State Initialization: The server initializes the global model with the model parameter \( \theta_g \)
\For{\( t = 0, \ldots, T-1 \)}
    \State \( S_t \gets \) Sample a subset of clients
    \State \( \theta_i^t \gets \theta_g^t \) 
    \For{each client \( i \in S_t \) in parallel}
        \State \( \theta_i^{t+1} \gets \) ClientUpdate(\( i, \theta_i^t, t \)) \Comment{Update client \( i \) and compute gradient norm}
    \EndFor
    \If{\( t = 0 \)} \Comment{Only execute at the initial training round}
    \For{each client \( i \in S_t \) in parallel}
    \State \colorbox{colorgreen}{\( g_i \gets \) GradNorm(\( i, \theta_i^t, t \))} \Comment{Calculate gradient norm for each client \( i \)}
    \EndFor
    \State \colorbox{colorgreen}{\( \mathcal{C}_{clean}, \mathcal{C}_{noisy} \gets C(\{g_i\}) \)} \Comment{Clustering client \( i \) into \emph{clean} or \emph{noisy} based on $g_i$}
    \State \colorbox{colorgreen}{ \( \alpha_t, \beta_t \gets W( \mathcal{C}_{clean}, \mathcal{C}_{noisy}, i) \)} \Comment{Assign weighting factors to each cluster}
    \Else
    \State \colorbox{colorgreen}{\( \alpha_t \gets \alpha_{t-1} \) and \( \beta_t \gets \beta_{t-1} \)}
    \EndIf
    
    \State \colorbox{colorred}{\( \theta_g^{t+1} \gets \theta_g^t - \eta \sum_{i \in S_t} \mathbf{w}_i \nabla \ell(\theta_i^t) \)} \hfill (\textbf{FedAvg})
    \State \colorbox{colorgreen}{\( \theta_g^{t+1} \gets \theta_g^t - \eta \left( \sum_{i \in n} \mathbf{w}_i \alpha_t \nabla \ell(\theta_i^t) + \sum_{j \in m} \mathbf{w}_j \beta_t \nabla \tilde{\ell}(\theta_j^t) \right) \)} \hfill (\textbf{FedAvg+\framework})
\EndFor
\Statex \textbf{Client executes:}
\State Local update: \textbf{ClientUpdate}(\( i, \theta_i, t \)) \Comment{Client \( i \) executes update at round \( t \)}
    \For{\( e = 0, \ldots, E-1 \)}
        \State Update \( \theta_i \)
    \EndFor
    \State \Return \( \theta_i \)
    \State \colorbox{colorgreen}{Compute gradient norm: \textbf{GradNorm}(\( i, \theta_i, t \))}
    \If{\( t = 0 \)} 
    \State \colorbox{colorgreen}{\(g_i \gets \| \nabla f(\theta_i) \| \)} \Comment{Compute gradient norm at the beginning round \( t_0 \)}
    \EndIf
    \State \Return \( g_i \)
\end{algorithmic}
\label{algorithm}
\end{algorithm*}
\section{Experiments}
\label{sec:exp}
\vspace{-0.1cm}

\subsection{Datasets, Models, and Configuration}
\label{sec:exp_data}
\textbf{Datasets.} To demonstrate the efficacy of \texttt{FedNS}, we develop the noisy datasets benchmark based on \cite{hendrycks2019robustness}. we consider the image classification task and test our approach across various datasets, including CIFAR10/100 \cite{krizhevsky2009learning}, Path-MNIST \cite{medmnistv2}, Fashion-MNIST \cite{xiao2017/online}, EuroSAT \cite{helber2019eurosat}, and Tiny-ImageNet \cite{le2015tiny}. These datasets are further applied data corruptions that are defined in Section \ref{sec:definition} to create the noisy input for FL. Specifically, we categorize data corruption into distortion and patch-based noise. In practice, distortion corruption consists of noise addition (e.g., Gaussian noise, shot noise, impulse noise, pixelated noise), blur (e.g., defocus blur, glass blur, motion blur, Gaussian blur), photometric distortions (e.g., brightness, contrast, saturate), synthetic distortions (e.g., snow, fog, elastic, spatter), and compression artifact (e.g., JPEG compression). Moreover, we further evaluate our method on patch-based obfuscation noise, including black patch, Gaussian noise patch, and StyleGAN-based patch \cite{baradad2021learning}.

\textbf{Baselines and Implementation.} We apply \framework~in conjunction with four widely used FL strategies to evaluate its effectiveness, namely FedAvg \cite{mcmahan2017communication}, FedProx \cite{li2020federated}, FedTrimmedAvg \cite{yin2018byzantine}, and FedNova \cite{wang2020tackling}. We employ ResNet-18 as the main model architecture, utilizing the mini-batch SGD as the universal local optimizer for all FL strategies. Moreover, we further evaluate \framework~on a different neural architecture ConvMixer-256/8 in Table \ref{tab:ConvMixer-exp}. The optimizer is characterized by a learning rate of $0.01$, an SGD momentum of $0.9$, and the weight decay is set to $10^{-4}$. We set the number of local training epochs to $5$ and the global communication rounds to $150$ across all datasets. 
\textcolor{black}{We consider a setup with $N = 20$ clients for our experiments unless mentioned otherwise. This choice aligns with standard practices in FL and accommodates our computational limitations.} The training set is distributed under both IID and non-IID settings. For the main experiments, we divide the client set into $15$ noisy clients and $5$ clean clients, with full client participation $r_p = 1.0$. \textcolor{black}{We compute $L_1$-norm of the gradient of last layer for noisy client detection in all the cases, see Table~\ref{Ablation: grad_norm} for comparison between the effectiveness of $L_1$ and $L_2$ norms, as well as the impact of batch size on the detection performance. We conducted all experiments on NVIDIA A10 GPUs.}

\begin{table*}[!t]
\centering
\caption{\small{Comparison of top-1 accuracy across datasets in IID and Non-IID settings.  We evaluate the performance of \framework~with various federated aggregation methods for learning under the noisy environment.}}
\label{tab:Main_Result_1}
\resizebox{\textwidth}{!}{%
\begin{tabular}{@{}lcccccccccccc@{}}
\toprule
\multirow{2}{*}{\textbf{Methods}} & \multicolumn{2}{c}{\textbf{CIFAR-10}} & \multicolumn{2}{c}{\textbf{CIFAR-100}} & \multicolumn{2}{c}{\textbf{PathMNIST}} & \multicolumn{2}{c}{\textbf{FMNIST}} & \multicolumn{2}{c}{\textbf{EuroSAT}} & \multicolumn{2}{c}{\textbf{Tiny-ImageNet}} \\
\cmidrule(r){2-3} \cmidrule(r){4-5} \cmidrule(r){6-7} \cmidrule(r){8-9} \cmidrule(r){10-11} \cmidrule(r){12-13}
& \textbf{IID} & \textbf{Non-IID} & \textbf{IID} & \textbf{Non-IID} & \textbf{IID} & \textbf{Non-IID} & \textbf{IID} & \textbf{Non-IID} & \textbf{IID} & \textbf{Non-IID} & \textbf{IID} & \textbf{Non-IID} \\
\midrule
FedAvg\cite{mcmahan2017communication} & 78.62\% & 73.51\% & 44.58\% & 42.10\% & 54.80\% & 52.14\% & 88.14\% & 84.67\% & 67.39\% & 75.06\% & 24.32\% & 22.90\% \\
+ \ours (Ours) & \textbf{81.67\%} & \textbf{78.44\%} & \textbf{48.14\%} & \textbf{45.94\%} & \textbf{63.89\%} & \textbf{62.92\%} & \textbf{89.61\%} & \textbf{88.53\%} & \textbf{78.22\%} & \textbf{80.12\%} & \textbf{27.85\%} & \textbf{25.93\%} \\
\midrule
FedProx\cite{li2020federated} & 79.89\% & 78.13\% & 46.75\% & 45.17\% & 57.28\% & 56.27\% & 87.15\% & 86.96\% & 70.83\% & 76.64\% & 24.90\% & 23.76\% \\
+ \ours (Ours) & \textbf{82.31\%} & \textbf{81.18\%} & \textbf{48.27\%} & \textbf{46.80\%} & \textbf{60.18\%} & \textbf{63.11\%} & \textbf{89.12\%} & \textbf{87.48\%} & \textbf{76.94\%} & \textbf{81.20\%} & \textbf{26.48\%} & \textbf{25.98\%} \\
\midrule
FedTrimmedAvg\cite{yin2018byzantine} & 78.92\% & 77.24\% & 41.81\% & 41.25\% & 56.34\% & 54.50\% & 90.09\% & 89.95\% & 68.30\% & 74.39\%  & 16.97\% & 15.48\% \\
+ \ours (Ours) & \textbf{82.63\%} & \textbf{82.47\%} & \textbf{49.11\%} & \textbf{48.32\%} & \textbf{64.27\%} & \textbf{63.04\%} & \textbf{90.29\%} & \textbf{91.57\%} & \textbf{83.81\%} & \textbf{80.50\%} & \textbf{29.43\%} & \textbf{27.46\%} \\
\midrule
FedNova\cite{wang2020tackling} & 81.45\% & 82.16\% & 49.48\% & 48.24\% & 55.36\% & 51.04\% & \textbf{90.65\%} & 89.68\% & 73.54\% & 66.29\% & 28.62\% & 27.24\% \\
+ \ours (Ours) & \textbf{88.65\%} & \textbf{88.34\%} & \textbf{59.19\%} & \textbf{59.17\%} & \textbf{80.82\%} & \textbf{81.89\%} & 90.57\% & \textbf{91.50\%} & \textbf{93.31\%} & \textbf{92.70\%} & \textbf{48.50\%} & \textbf{46.16\%} \\
\bottomrule
\end{tabular}%
}
\end{table*}

\begin{figure}[t]
 \centering
 \begin{subfigure}[b]{0.24\textwidth}
     \centering
     \includegraphics[width=\textwidth]{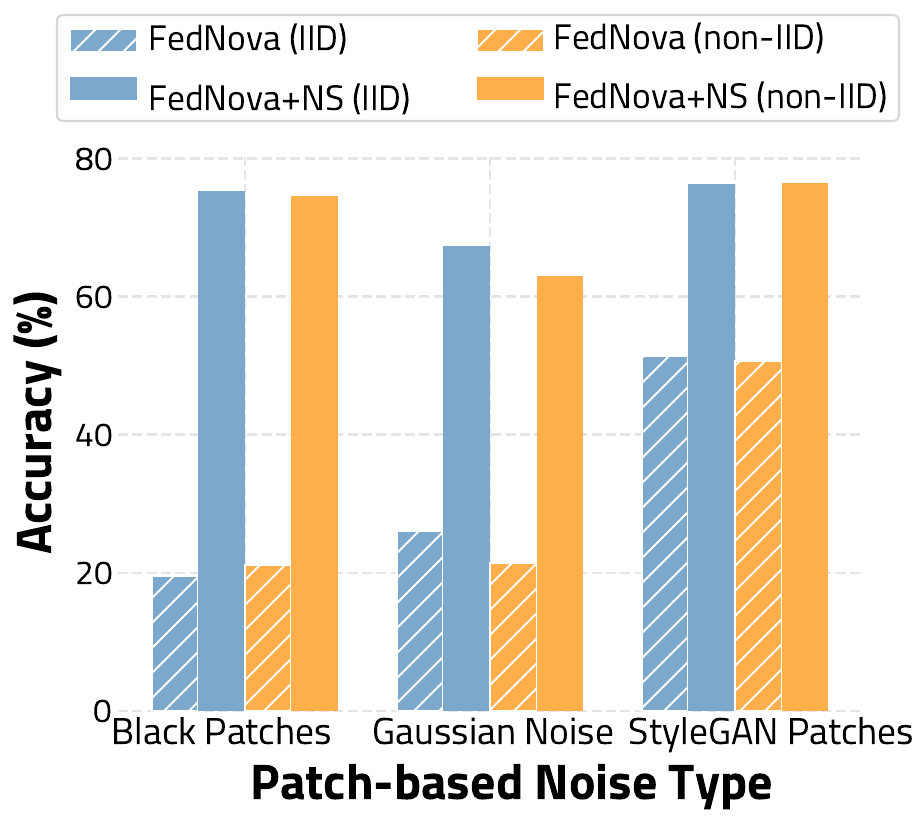}
     \caption{CIFAR-10}
     \label{subfig:patch-cifar10}
 \end{subfigure}
 \begin{subfigure}[b]{0.24\textwidth}
     \centering
     \includegraphics[width=\textwidth]{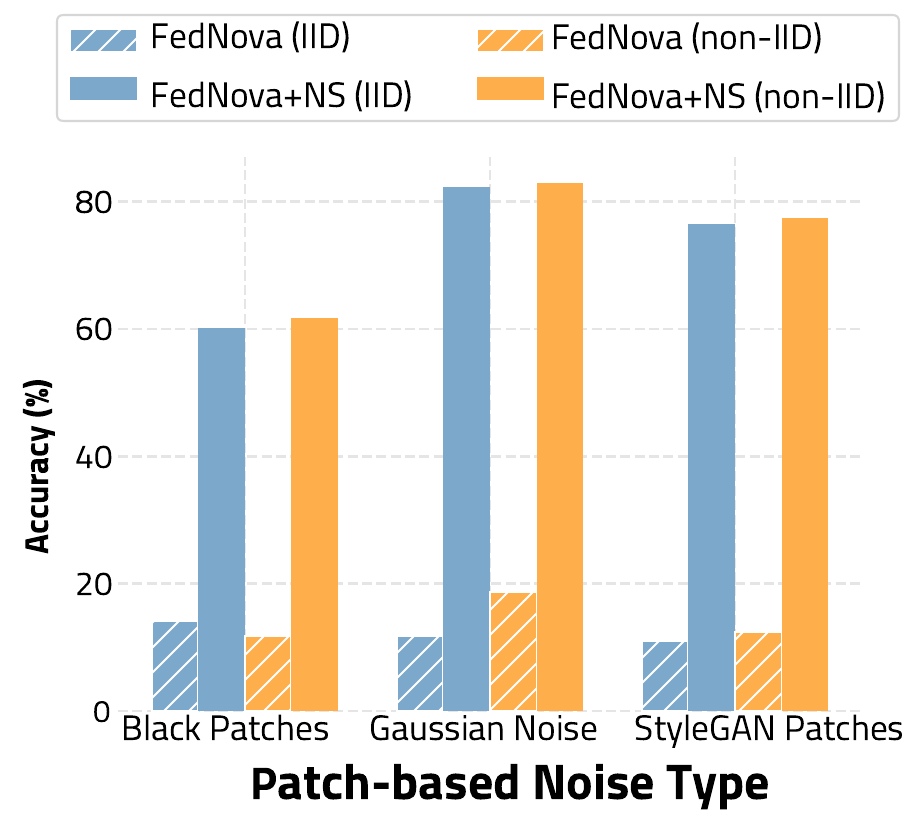}
     \caption{Path-MNIST}
     \label{subfig:patch-pathmnist}
 \end{subfigure}
    \caption{\small{Comparative results of \framework~in dealing with patch-based noise types on CIFAR-10 and Patch-MNIST. For consistency, we utilize the same noise setup as in Table \ref{tab:Main_Result_1} and conduct the experiments on both IID and non-IID settings.}}
    \label{fig:patch-based-noise}
    \vspace{-6mm}
\end{figure}

\begin{figure*}[h]
 \centering
    \includegraphics[width=0.8\textwidth]{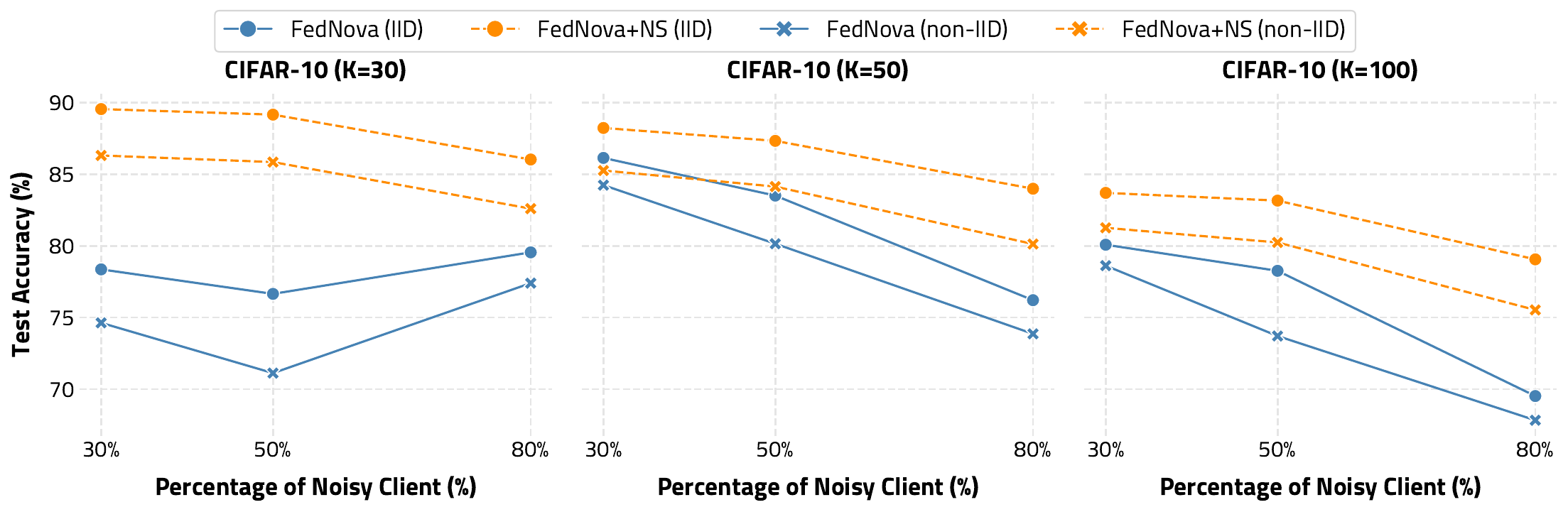}
    \caption{\small{Impact of the number of clients on model performance under different levels of client data distribution noise with FedNova.}}
    \label{ablation:client_amount}
    \vspace{-5mm}
\end{figure*}

\subsection{Results and Ablation Studies}
\label{sec:config}
\textbf{Participation of noisy clients deteriorates the performance of the global model.} To validate the efficacy of our proposed method, we first conduct an experiment for model training with clean and noisy input across all the datasets and utilize the same noise configuration for our further empirical evaluation. With this, we aim to evaluate the upper-bound performance that can be achieved when learning from a mixture of noisy and clean clients. Table \ref{tab:NoiseImpact} presents the comparative results of average accuracy for all considered datasets. We focus on three specific distortions (i.e., defocus blur, Gaussian blur, contrast) due to their significant impact on degrading the model's generalization capability to simulate the worst case in noisy FL.
For the generation of distorted data used in the experiments summarized in Table \ref{tab:Main_Result_1}, each noisy sample was produced by randomly selecting a distortion type with the configuration characterized by the noise severity level $\xi = high$. We set the noise level to $\emph{NL}_m = 100\%$ for every client $m$ for the experiments of both Table \ref{tab:Main_Result_1} and Figure \ref{fig:patch-based-noise}. We see the participation of noisy clients leads to a significant degradation in the model's generalization capability across all tasks, indicating the detrimental impact of noisy data in the FL environment. Furthermore, due to the inherent lack of visibility into the data from these federated clients, the resultant global model tends to be of low quality and it may become challenging in a real-world setting to identify the underlying reasons for its poor performance.

\textbf{FedNS significantly improves standard federated aggregation methods.} We investigate the robustness of our proposed method by applying \texttt{FedNS} on six image datasets with different settings under the noisy scenario. As shown in Table \ref{tab:Main_Result_1}, the performance of all aggregation methods exhibits a general trend of improvement by simply plugging FedNS to the considered strategies. In particular, we consider the worst-case with heterogeneous data setting in Table \ref{tab:Main_Result_1}, where 15 out of 20 noisy clients participate in the federated training with high noise severity and $100$\% noise level. Adding \texttt{FedNS} to FL strategies yields better overall performance among all the datasets, especially for some vulnerable datasets (e.g., Path-MNIST) that are sensitive to data corruption. Additionally, we demonstrate the efficacy of \texttt{FedNS} in dealing with patch-based noise on CIFAR-10 and Path-MNIST as presented in Figure \ref{fig:patch-based-noise}. From Figure \ref{fig:patch-based-noise}, we observe that FedNS consistently boosts the performance of the aggregation method across all the datasets and settings. This further shows that FedNS is capable of handling various types of noises a model can encounter in a real-world setting. 

\textbf{Impact of noisy client distributions on performance across varying numbers of clients.} We evaluate FedNS under the influence of noise client percentages with different amounts of clients at scale, where $K$ denotes the number of clients. We select $K$ from $\{ 30, 50, 100 \}$ and set the noisy client percentage as $30$\%, $50$\%, $80$\% , respectively. As shown in the figure, the results exhibit a consistent pattern of variation in model performance across different scenarios. Specifically, as the percentage of noisy clients increases, there is a gradual decline in test accuracy for both IID and non-IID distributions. However, FedNova+NS (IID) consistently outperforms the other methods, maintaining higher accuracy levels across all settings. This demonstrates the robustness and effectiveness of the FedNova+NS method in mitigating the adverse effects of noisy clients, especially as the number of clients increases.

\begin{table*}[t]
\centering
\caption{\small{Top-1 accuracy of the ConvMixer-256/8 model using FedNova aggregation strategy in a high-noise environment.}}
\label{tab:ConvMixer-exp}
\resizebox{1.0\textwidth}{!}{%
\begin{tabular}{@{}lcccccccccccc@{}}
\toprule
\multirow{2}{*}{\textbf{Methods}} & \multicolumn{2}{c}{\textbf{CIFAR-10}} & \multicolumn{2}{c}{\textbf{CIFAR-100}} & \multicolumn{2}{c}{\textbf{PathMNIST}} & \multicolumn{2}{c}{\textbf{FMNIST}} & \multicolumn{2}{c}{\textbf{EuroSAT}} & \multicolumn{2}{c}{\textbf{Tiny-ImageNet}} \\
\cmidrule(r){2-3} \cmidrule(r){4-5} \cmidrule(r){6-7} \cmidrule(r){8-9} \cmidrule(r){10-11} \cmidrule(r){12-13} & \textbf{IID} & \textbf{Non-IID} & \textbf{IID} & \textbf{Non-IID} & \textbf{IID} & \textbf{Non-IID} & \textbf{IID} & \textbf{Non-IID} & \textbf{IID} & \textbf{Non-IID} & \textbf{IID} & \textbf{Non-IID} \\
\midrule
FedNova & 71.76\% & 64.50\% & 37.31\% & 36.23\% & 54.15\% & 49.75\% & 84.43\% & 86.31\% & 61.70\% & 56.28\% & 21.04\% & 19.43\% \\
+ \ours (Ours) & \textbf{77.05\%} & \textbf{70.51\%} & \textbf{44.17\%} & \textbf{43.07\%} & \textbf{79.76\%} & \textbf{80.82\%} & \textbf{85.68\%} & \textbf{86.71\%} & \textbf{88.59\%} & \textbf{87.80\%} & \textbf{33.66\%} & \textbf{31.80\%} \\

\bottomrule
\end{tabular}%
}
\vspace{-2mm}
\end{table*}

\begin{figure*}[t]
 \centering
 \begin{subfigure}[t]{0.45\textwidth}
     \centering
     \includegraphics[width=\textwidth]{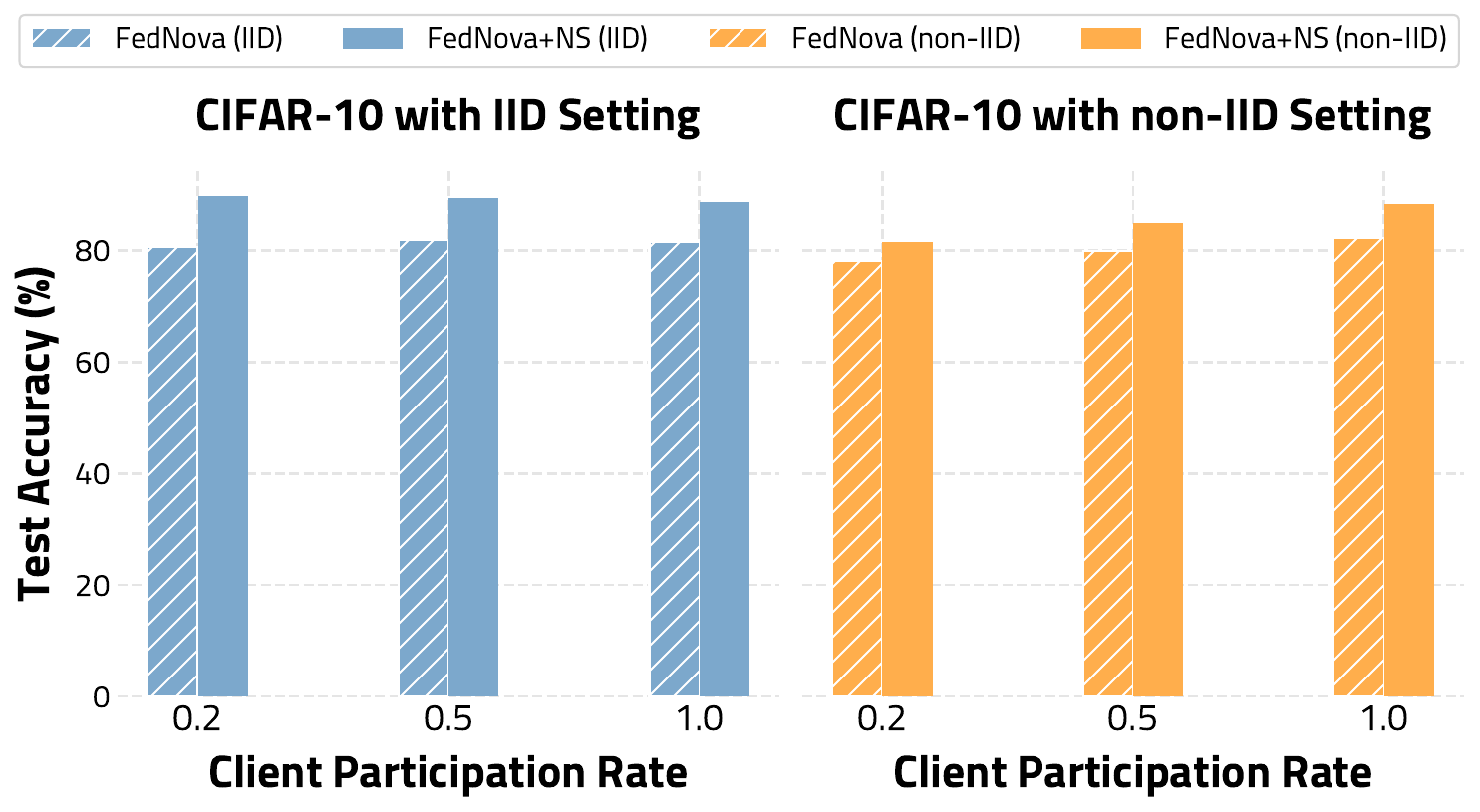}
     \caption{CIFAR-10}
     \label{sub:cifar100}
 \end{subfigure}
 \hfill
 \begin{subfigure}[t]{0.45\textwidth}
     \centering
     \includegraphics[width=\textwidth]{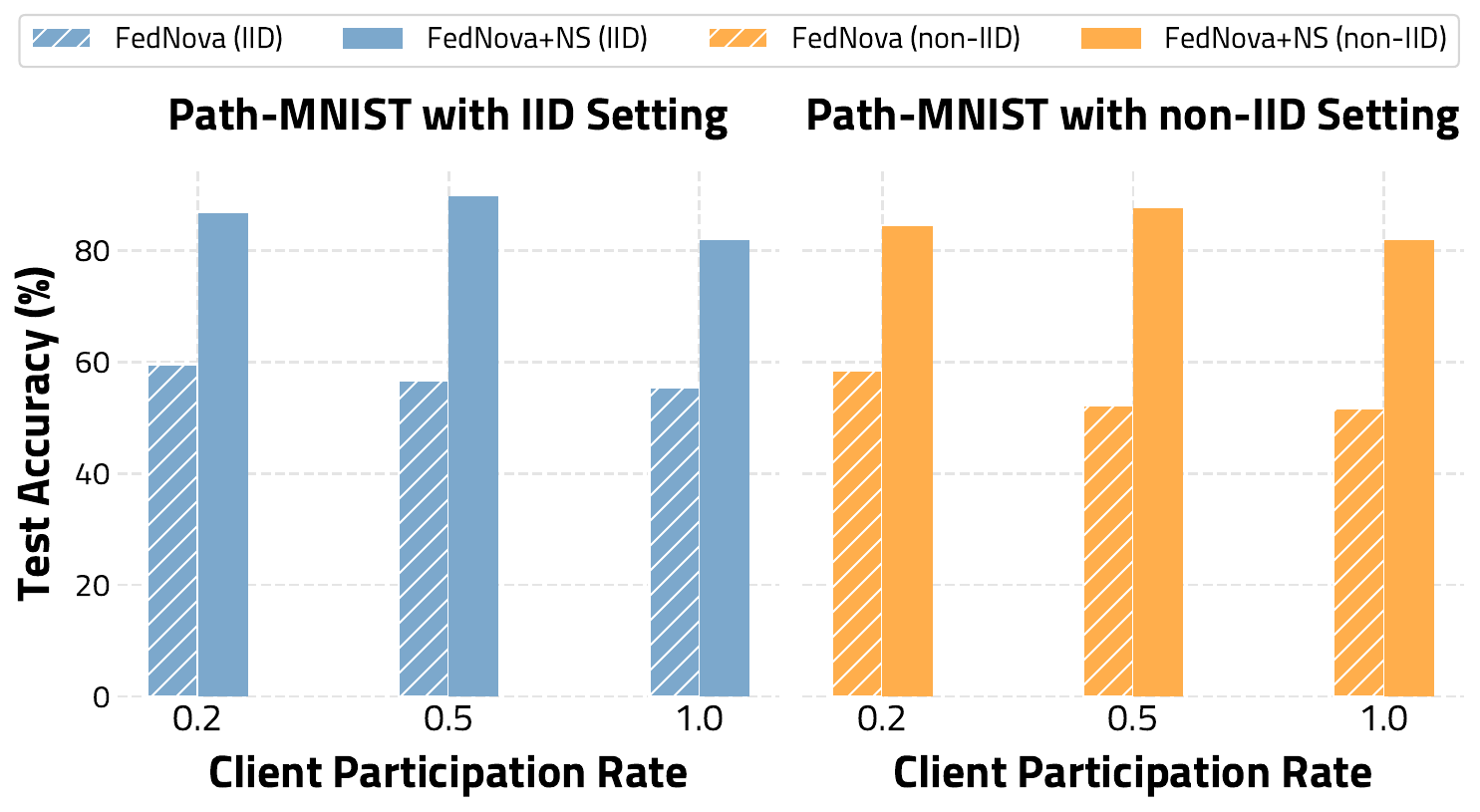}
     \caption{Path-MNIST}
     \label{sub:pathmnist}
 \end{subfigure}
    \caption{\small{Ablation on the impact of different client participation rate ($r_p$). Evaluation on the CIFAR-10 and Path-MNIST datasets, with varying $r_p$ selections. The results compare model generalization performance with and without \framework~across different experimental setups.}}
    \label{ablation:client_pool}
    \vspace{-5mm}
\end{figure*}

\textbf{Impact of Client Participation Rate ($r_p$).} Next, we examine how the client participation rate ($r_p$) affects the performance of  \framework. We conduct experiments on CIFAR-10 and Path-MNIST, maintaining a fixed noise configuration across all settings. We vary $r_p$ to simulate different device participation patterns, choosing values from $r_p \in {0.2, 0.5, 1.0}$, where $r_p = 1.0$ denotes full client participation. In practice, the server ensures full client participation during the first training round to identify noisy clients. Figure \ref{ablation:client_pool} provides the results. We observe that incorporating  \framework~significantly enhances performance across all client participation patterns. Our experiments suggest that \framework~remains robust to variations in client availability and data quality.

\textbf{Robustness of \framework~on mixed noise conditions.} Next, we investigate the robustness of \framework~under complex noise conditions that involve a combination of different noise types, including the distortions and patch-based noises as described in Section \ref{sec:exp_data}. Table \ref{tab:mixed_all_noises} shows the performance gains achieved by \framework~are particularly significant in high-variance datasets such as Path-MNIST and Tiny-ImageNet, where FedNova+NS demonstrates substantial improvements over standard aggregation. Notably, \framework~consistently enhances the global model's performance across all evaluated datasets, showcasing its robustness in handling intricate noise scenarios. Our results highlight the effectiveness of \framework~in mitigating the impact of complex noise conditions, where decentralized data may be subject to various types of distortions at the same time.

\begin{table}[h]
\centering
\caption{\small{Top-1 accuracy of the FedNova aggregation method (+\framework) in an IID setting on selected datasets when noisy clients' data is a mixture of all noise types.}}
\label{tab:mixed_all_noises}
\resizebox{\columnwidth}{!}{
    \begin{tabular}{@{}lccc@{}}
    \toprule
    \textbf{Noise Setting} & \textbf{Clean} & \multicolumn{2}{c}{\textbf{Noisy}} \\
    \cmidrule(lr){2-2} \cmidrule(lr){3-4}
    \textbf{Methods} & \textbf{FedNova} & \textbf{FedNova} & \textbf{FedNova+NS} \\
    \midrule
    CIFAR-10 & 90.14\% & 80.10\% & 85.83\% (\textbf{$\uparrow$ 5.73\%}) \\
    \midrule
    CIFAR-100 & 64.79\% & 45.28\% & 51.47\% (\textbf{$\uparrow$ 6.19\%}) \\
    \midrule
    Path-MNIST & 92.34\% & 65.19\% & 87.81\% (\textbf{$\uparrow$ 22.62\%}) \\
    \midrule
    Tiny-ImageNet & 53.26\% & 29.26\% & 43.30\% (\textbf{$\uparrow$ 14.04\%}) \\
    \bottomrule 
    \end{tabular}
}
\end{table}

\textbf{Evaluation on Weight Factor($\beta$) of Noisy Client.} We further investigate the effect of weight factor $\beta$ on overall performance in FedNS. The weight factor $\beta$ controls the weights of aggregated noisy client local models, as described in Equation \ref{eq:global_update}. We considered various values of $\beta \in \{0, 0.1, 0.3, 0.5, 0.7, 1.0 \}$, where $\beta=0$ signifies the exclusion of noisy model weights, and $\beta=1.0$ indicates the direct aggregation of model weights. As shown in Table \ref{ablation: noisy_weight_factor}, the setting of $\beta=0.3$ yields the best performance, suggesting an optimal trade-off. Interestingly, the exclusion of noisy model weights ($\beta=0.0$) leads to a degradation in the generalization capability of the global model. This observation suggests that incorporating mitigated noisy data enhances the robustness of the global model.


\begin{table}[h]
    \centering
    \caption{\small{Ablation on the weight factor $\beta$ for noisy clients. We evaluate this parameter on the CIFAR-10 dataset with IID and non-IID settings, and the weight factor of clean clients $\alpha$ is set to $2.0$.}}
    \label{ablation: noisy_weight_factor}
        \centering
        \resizebox{\columnwidth}{!}{
        \begin{tabular}{@{}l*{6}{c}@{}}
        \toprule
        \multirow{2}{*}{\textbf{Setting}} & \multicolumn{6}{c}{\textbf{Weight Factor $\beta$ of Noisy Client}} \\
        \cmidrule(l){2-7}
        & $\mathbf{0.0}$ & $\mathbf{0.1}$ & $\mathbf{0.3}$ & $\mathbf{0.5}$ & $\mathbf{0.7}$ & $\mathbf{1.0}$ \\
        \midrule
        IID & 82.82 & 87.17 & 88.65 & 87.53 & 86.28 & 84.33 \\
        non-IID & 75.31 & 82.49 & 88.34 & 84.05 & 83.44 & 81.27 \\
        \bottomrule
        \end{tabular}
        }
        \vspace{-2mm}
\end{table}

\textbf{FedNS on Real-world Human Annotation Errors.} In this experiment, we extend our investigation to assess the efficacy of FedNS in addressing real-world data quality issues, specifically human annotation errors. While this work focuses on mitigating data corruption in the input space, we identify that FedNS can effectively handle label noise - a well-studied problem in noisy FL. Label noise occurs when data labels are incorrectly assigned while the input features remain unaltered. Specifically, we assessed FedNS on CIFAR-10/100N, two benchmark datasets featuring real-world noisy labels resulting from human annotation errors \cite{wei2021learning}. We employed FedAvg and FedNova as baseline approaches, adhering to the training configuration outlined in Section \ref{sec:exp_data}. From Table \ref{tab:label_noise}, we observe that FedNS consistently improved the performance across all the experiments. Our findings demonstrate that FedNS not only excels in mitigating input space corruption but also shows promising results in handling label noise. The ability to address both input space corruption and label noise positions FedNS as a valuable tool for practitioners dealing with real-world datasets, where multiple types of data imperfections may coexist.

\begin{table}[h]
\centering
\caption{\small{Performance of FedNS on datasets with noisy labels. We evaluate FedNS on CIFAR-10/100N \cite{wei2021learning} datasets.}}
\label{tab:label_noise}
\resizebox{0.8\columnwidth}{!}{%
\begin{tabular}{@{}lcccc@{}}
\toprule
\multirow{2}{*}{\textbf{Methods}} & \multicolumn{2}{c}{\textbf{CIFAR-10N}} & \multicolumn{2}{c}{\textbf{CIFAR-100N}} \\
\cmidrule(r){2-3} \cmidrule(r){4-5} & \textbf{IID} & \textbf{Non-IID} & \textbf{IID} & \textbf{Non-IID} \\
\midrule
FedAvg & 76.06\% & 69.52\% & 52.61\% & 52.57\%  \\
+ \ours (Ours) & \textbf{78.87\%} & \textbf{71.14\%} & \textbf{54.06\%} & \textbf{53.85\%}\\
\midrule
FedNova & 76.06\% & 69.26\% & 53.32\% & 52.41\%\\
+ \ours (Ours) & \textbf{84.38\%} & \textbf{78.65\%} & \textbf{55.54\%} & \textbf{54.80\%}\\
\bottomrule
\end{tabular}%
}
\end{table}

\textbf{Impact of Initial-Round Client Participation on FedNS Performance.} In FL, the assumption that all clients must be warmed up and participate from the very first training round may not always hold true, as some clients may only participate in later rounds. To address this, we explore the scenario where only a subset of clients participates in the initial round. Instead of collecting gradient norms solely in the first round, we iteratively apply FedNS until all clients have been engaged. We evaluate this approach by setting the first-round participation rates to 10\%, 50\%, and 100\%, with 100\% representing full client participation. The results, presented in Table \ref{tab:initial_participation}, highlight the robustness of FedNS across varying levels of initial-round client participation.

\begin{table}[h]
\centering
\caption{\small{FedNS performance on CIFAR-10 and CIFAR-100: Classification accuracy with 10\%, 50\%, and 100\% client participation in the first round, i.e., partially warmed-up clients.}}
\label{tab:initial_participation}
\resizebox{0.5\textwidth}{!}{%
\begin{tabular}{@{}lcccccc@{}}
\toprule
\textbf{Methods} & \multicolumn{3}{c}{\textbf{CIFAR-10}} & \multicolumn{3}{c}{\textbf{CIFAR-100}} \\
\cmidrule(r){2-4} \cmidrule(r){5-7} \textbf{First-Round Participation} & \textbf{10\%} & \textbf{50\%} & \textbf{100\%} & \textbf{10\%} & \textbf{50\%} & \textbf{100\%} \\
\midrule
FedNova & 80.46\% & 81.14\% & 81.45\% & 48.63\% & 48.46\% & 49.48\% \\
+ \ours (Ours) & \textbf{86.13\%} & \textbf{86.76\%} & \textbf{88.65\%} & \textbf{54.13\%} & \textbf{54.27\%} & \textbf{59.19\%} \\
\bottomrule
\end{tabular}
}
\vspace{-1mm}
\end{table}

\textbf{FedNS on Different Noise Level ($\emph{NL}$) of Client Data.} In Table \ref{Ablation: noise_level}, we examine the performance of \framework~across various noise configurations. Specifically, we consider noise levels $\emph{NL} \in \{ 50\%, 80\%, 100\% \}$ and noise severity $\xi \in \{ Medium, High \}$. These results demonstrate that \framework~substantially enhances model generalization under high noise conditions, especially when clients' data is completely corrupted (i.e., $100$\% noise level and high severity). Conversely, in scenarios with milder noise where the impact on federated models is minimal, the implementation of \framework~does not negatively affect the generalization process. Hence, we limit our experiments to noise levels at or above 50\%, as noise levels below this threshold have negligible impact on model performance.

\begin{table}[ht]
\centering
\caption{\small{Ablation for different noise level and noise severity with IID setting on CIFAR-10/100 datasets. We employ FedNova as the aggregation method and ablate the noise level and noise severity.}}
\label{Ablation: noise_level}
\begin{subtable}[h]{0.49\textwidth}
    \centering
    \resizebox{\textwidth}{!}{
    \begin{tabular}{@{}lcccccc@{}}
    \toprule
    \textbf{Noise Level($\emph{NL}$)} & \multicolumn{2}{c}{$\emph{NL} = \mathbf{50}\textbf{\%}$} & \multicolumn{2}{c}{$\emph{NL} = \mathbf{80}\textbf{\%}$} & \multicolumn{2}{c}{$\emph{NL} = \mathbf{100}\textbf{\%}$} \\
    \cmidrule(r){2-3} \cmidrule(r){4-5} \cmidrule(r){6-7} \textbf{Noise Severity} & Medium & High & Medium & High & Medium & High \\
    \midrule
    FedNova  & 89.95\% & 89.02\% & 90.02\% & 88.90\% & 87.95\% & 81.45\% \\
    + \ours (Ours)  & \textbf{90.56\%} & \textbf{90.11\%} & \textbf{90.35\%} & \textbf{90.12\%} & \textbf{90.35\%} & \textbf{88.65\%} \\
    \bottomrule
    \end{tabular}
    }
    \vspace{1pt}
    \caption{CIFAR-10}
    \label{subtab:CIFAR10}
\end{subtable}
\hfill
\begin{subtable}[h]{0.49\textwidth}
    \centering
    \resizebox{\textwidth}{!}{
    \begin{tabular}{@{}lcccccc@{}}
    \toprule
    \textbf{Noise Level($\emph{NL}$)} & \multicolumn{2}{c}{$\emph{NL} = \mathbf{50}\textbf{\%}$} & \multicolumn{2}{c}{$\emph{NL} = \mathbf{80}\textbf{\%}$} & \multicolumn{2}{c}{$\emph{NL} = \mathbf{100}\textbf{\%}$} \\
    \cmidrule(r){2-3} \cmidrule(r){4-5} \cmidrule(r){6-7} \textbf{Noise Severity} & Medium & High & Medium & High & Medium & High \\
    \midrule
    FedNova & \textbf{65.51\%}  & 62.70\% & 64.18\% & 60.96\% & 60.85\% & 49.48\% \\
    + \ours (Ours) & 65.03\% & \textbf{62.83\%} & \textbf{65.17\%} & \textbf{62.14\%} & \textbf{62.76\%} & \textbf{59.19\%} \\
    \bottomrule
    \end{tabular}
    }
    \vspace{1pt}
    \caption{CIFAR-100}
    \label{subtab:CIFAR100}
\end{subtable}
\vspace{-5mm}
\end{table}

\textbf{Hyperparameters Selections for Noisy Clients Detection.} One of the essential components of \framework~is using gradient norms to identify noisy clients. We investigate several factors that may affect the performance of detecting noisy clients, specifically the selection of the $L_p$-norm and the batch size of gradient norms for clustering. This ablation experiment is performed on the CIFAR-10 and Path-MNIST datasets, as presented in Table \ref{Ablation: grad_norm}.

The comparison between the $L_1$-norm and $L_2$-norm shows that the $L_1$-norm generally outperforms higher-order norms. Moreover, we evaluate the impact of batch size by setting the mini-batch size to range from 1 to 128. Our experiments suggest that a larger batch size effectively reduces the variance of the gradient estimates. The results, shown in Table \ref{Ablation: grad_norm}, indicate that a batch size selection around $16 \sim 64$ yields better performance across all settings.

\begin{table}[ht]
    \centering
    \caption{\small{Hyperparameter impact on noisy client detection: CIFAR-10 and Path-MNIST, FedNova, 5 clean + 15 noisy clients (100\% noise), IID, full participation in first round, 5 local epochs. Values show noisy client detection accuracy.}}
    \begin{subtable}[h]{0.24\textwidth}
        \centering
        \resizebox{\textwidth}{!}{
        \begin{tabular}{@{}l*{5}{c}@{}}
        \toprule
        \multirow{2}{*}{\textbf{$L_p$-Norm}} & \multicolumn{5}{c}{\textbf{Batch Size of Gradient Norm}} \\
        \cmidrule(l){2-6}
        & $\mathbf{1}$ & $\mathbf{16}$ & $\mathbf{32}$ & $\mathbf{64}$ & $\mathbf{128}$ \\
        \midrule
        $L_1$-Norm & 0.85 & 1.00 & 1.00 & 0.95 & 0.90 \\
        $L_2$-Norm & 0.90 & 0.95 & 0.90 & 0.90 & 0.95 \\
        \bottomrule
        \end{tabular}
        }
        \vspace{2pt}
        \caption{CIFAR-10}
        \label{subtab:norm-CIFAR10}
    \end{subtable}
    \hfill
    \begin{subtable}[h]{0.24\textwidth}
        \centering
        \resizebox{\textwidth}{!}{
        \begin{tabular}{@{}l*{5}{c}@{}}
        \toprule
        \multirow{2}{*}{\textbf{$L_p$-Norm}} & \multicolumn{5}{c}{\textbf{Batch Size of Gradient Norm}} \\
        \cmidrule(l){2-6}
        & $\mathbf{1}$ & $\mathbf{16}$ & $\mathbf{32}$ & $\mathbf{64}$ & $\mathbf{128}$ \\
        \midrule
        $L_1$-Norm & 0.75 & 1.00 & 1.00 & 1.00 & 0.90 \\
        $L_2$-Norm & 0.60 & 0.75 & 0.85 & 0.90 & 0.90 \\
        \bottomrule
        \end{tabular}
        }
        \vspace{2pt}
        \caption{Path-MNIST}
        \label{subtab:norm-CIFAR100}
    \end{subtable}
    \label{Ablation: grad_norm}
    \vspace{-4mm}
\end{table}

\textbf{Evaluating \framework~with alternative model architecture.} In this experiment, we investigate the robustness of \framework~when applied to a different model architecture. We employ the ConvMixer-256/8 \cite{trockman2022patches} model and train it using FedNova on a range of datasets. The noise configuration remains consistent with the details provided in Section \ref{sec:config}, and we evaluate the model's performance under both IID and non-IID settings. As shown in Table \ref{tab:ConvMixer-exp}, the federated aggregation method, when paired with \framework, achieves enhanced performance across all considered datasets. These improvements, observed consistently across tasks, demonstrate the adaptability and effectiveness of \framework~when integrated with alternative architectural paradigms. The results highlight the versatility of our approach in enhancing the performance of different federated model regardless of the neural architecture, further emphasizing its potential as a robust and flexible method for FL.

\vspace{-0.1cm}
\section{Conclusion}
\label{sec:conclusion}
\vspace{-0.2cm}
In this paper, we study the training of the federated neural networks within a noisy environment in which the input space of client data is corrupted. We propose to utilize the gradient norm of local training updates to discover the noisy clients. We then propose \framework, an effective method designed to integrate with diverse FL aggregation methods to mitigate the impact of model updates from noisy clients. Our experimental results on several benchmark datasets show that \framework~is robust against data perturbation, which can significantly boost the generalization of federated models. We further present extensive ablation studies to provide a better understanding of \framework. Our findings highlight the importance for FL approaches that prioritize robustness when training models on decentralized data. By focusing on this aspect, we can enhance the reliability and stability of FL in a real-world setting, where data quality may vary significantly across participating clients. \textcolor{black}{Future work will investigate~\framework's efficacy in the language modeling domain and learning from multi-modal noisy data in a decentralized setting.}

\bibliography{IEEEtran}

\begin{thebibliography}{10}
\providecommand{\url}[1]{#1}
\csname url@samestyle\endcsname
\providecommand{\newblock}{\relax}
\providecommand{\bibinfo}[2]{#2}
\providecommand{\BIBentrySTDinterwordspacing}{\spaceskip=0pt\relax}
\providecommand{\BIBentryALTinterwordstretchfactor}{4}
\providecommand{\BIBentryALTinterwordspacing}{\spaceskip=\fontdimen2\font plus
\BIBentryALTinterwordstretchfactor\fontdimen3\font minus \fontdimen4\font\relax}
\providecommand{\BIBforeignlanguage}[2]{{%
\expandafter\ifx\csname l@#1\endcsname\relax
\typeout{** WARNING: IEEEtran.bst: No hyphenation pattern has been}%
\typeout{** loaded for the language `#1'. Using the pattern for}%
\typeout{** the default language instead.}%
\else
\language=\csname l@#1\endcsname
\fi
#2}}
\providecommand{\BIBdecl}{\relax}
\BIBdecl

\bibitem{konevcny2016federated}
J.~Kone{\v{c}}n{\`y}, H.~B. McMahan, F.~X. Yu, P.~Richt{\'a}rik, A.~T. Suresh, and D.~Bacon, ``Federated learning: Strategies for improving communication efficiency,'' \emph{arXiv preprint arXiv:1610.05492}, 2016.

\bibitem{mcmahan2017communication}
B.~McMahan, E.~Moore, D.~Ramage, S.~Hampson, and B.~A. y~Arcas, ``Communication-efficient learning of deep networks from decentralized data,'' in \emph{Artificial intelligence and statistics}.\hskip 1em plus 0.5em minus 0.4em\relax PMLR, 2017.

\bibitem{gudivada2017data}
V.~Gudivada, A.~Apon, and J.~Ding, ``Data quality considerations for big data and machine learning: Going beyond data cleaning and transformations,'' \emph{International Journal on Advances in Software}, 2017.

\bibitem{budach2022effects}
L.~Budach, M.~Feuerpfeil, N.~Ihde, A.~Nathansen, N.~Noack, H.~Patzlaff, F.~Naumann, and H.~Harmouch, ``The effects of data quality on machine learning performance,'' \emph{arXiv preprint arXiv:2207.14529}, 2022.

\bibitem{kairouz2021advances}
P.~Kairouz, H.~B. McMahan, B.~Avent, A.~Bellet, M.~Bennis, A.~N. Bhagoji, K.~Bonawitz, Z.~Charles, G.~Cormode, R.~Cummings \emph{et~al.}, ``Advances and open problems in federated learning,'' \emph{Foundations and trends{\textregistered} in machine learning}, 2021.

\bibitem{dodge2017study}
S.~Dodge and L.~Karam, ``A study and comparison of human and deep learning recognition performance under visual distortions,'' in \emph{2017 26th international conference on computer communication and networks (ICCCN)}.\hskip 1em plus 0.5em minus 0.4em\relax IEEE, 2017.

\bibitem{gupta2021data}
N.~Gupta, S.~Mujumdar, H.~Patel, S.~Masuda, N.~Panwar, S.~Bandyopadhyay, S.~Mehta, S.~Guttula, S.~Afzal, R.~Sharma~Mittal \emph{et~al.}, ``Data quality for machine learning tasks,'' in \emph{Proceedings of the 27th ACM SIGKDD conference on knowledge discovery \& data mining}, 2021.

\bibitem{huang2021importance}
R.~Huang, A.~Geng, and Y.~Li, ``On the importance of gradients for detecting distributional shifts in the wild,'' \emph{Advances in Neural Information Processing Systems}, 2021.

\bibitem{li2020federated}
T.~Li, A.~K. Sahu, M.~Zaheer, M.~Sanjabi, A.~Talwalkar, and V.~Smith, ``Federated optimization in heterogeneous networks,'' \emph{Proceedings of Machine learning and systems}, 2020.

\bibitem{yin2018byzantine}
D.~Yin, Y.~Chen, R.~Kannan, and P.~Bartlett, ``Byzantine-robust distributed learning: Towards optimal statistical rates,'' in \emph{International Conference on Machine Learning}.\hskip 1em plus 0.5em minus 0.4em\relax Pmlr, 2018.

\bibitem{wang2020tackling}
J.~Wang, Q.~Liu, H.~Liang, G.~Joshi, and H.~V. Poor, ``Tackling the objective inconsistency problem in heterogeneous federated optimization,'' \emph{Advances in neural information processing systems}, 2020.

\bibitem{dodge2016understanding}
S.~Dodge and L.~Karam, ``Understanding how image quality affects deep neural networks,'' in \emph{2016 eighth international conference on quality of multimedia experience (QoMEX)}.\hskip 1em plus 0.5em minus 0.4em\relax IEEE, 2016.

\bibitem{NEURIPS2018_0937fb58}
R.~Geirhos, C.~R.~M. Temme, J.~Rauber, H.~H. Sch\"{u}tt, M.~Bethge, and F.~A. Wichmann, ``Generalisation in humans and deep neural networks,'' in \emph{Advances in Neural Information Processing Systems}, S.~Bengio, H.~Wallach, H.~Larochelle, K.~Grauman, N.~Cesa-Bianchi, and R.~Garnett, Eds., 2018.

\bibitem{borkar2019deepcorrect}
T.~S. Borkar and L.~J. Karam, ``Deepcorrect: Correcting dnn models against image distortions,'' \emph{IEEE Transactions on Image Processing}, 2019.

\bibitem{liang2020deep}
D.~Liang, X.~Gao, W.~Lu, and L.~He, ``Deep multi-label learning for image distortion identification,'' \emph{Signal processing}, 2020.

\bibitem{zheng2016improving}
S.~Zheng, Y.~Song, T.~Leung, and I.~Goodfellow, ``Improving the robustness of deep neural networks via stability training,'' in \emph{Proceedings of the ieee conference on computer vision and pattern recognition}, 2016.

\bibitem{zhou2017classification}
Y.~Zhou, S.~Song, and N.-M. Cheung, ``On classification of distorted images with deep convolutional neural networks,'' in \emph{2017 IEEE International conference on acoustics, speech and signal processing (ICASSP)}.\hskip 1em plus 0.5em minus 0.4em\relax IEEE, 2017.

\bibitem{patrini2017making}
G.~Patrini, A.~Rozza, A.~Krishna~Menon, R.~Nock, and L.~Qu, ``Making deep neural networks robust to label noise: A loss correction approach,'' in \emph{Proceedings of the IEEE conference on computer vision and pattern recognition}, 2017.

\bibitem{li2019learning}
J.~Li, Y.~Wong, Q.~Zhao, and M.~S. Kankanhalli, ``Learning to learn from noisy labeled data,'' in \emph{Proceedings of the IEEE/CVF conference on computer vision and pattern recognition}, 2019.

\bibitem{ghosh2017robust}
A.~Ghosh, H.~Kumar, and P.~S. Sastry, ``Robust loss functions under label noise for deep neural networks,'' in \emph{Proceedings of the AAAI conference on artificial intelligence}, 2017.

\bibitem{han2020survey}
B.~Han, Q.~Yao, T.~Liu, G.~Niu, I.~W. Tsang, J.~T. Kwok, and M.~Sugiyama, ``A survey of label-noise representation learning: Past, present and future,'' \emph{arXiv preprint arXiv:2011.04406}, 2020.

\bibitem{chen2020dealing}
Y.~Chen, X.~Yang, X.~Qin, H.~Yu, P.~Chan, and Z.~Shen, ``Dealing with label quality disparity in federated learning,'' \emph{Federated Learning: Privacy and Incentive}, 2020.

\bibitem{yang2021client}
M.~Yang, H.~Qian, X.~Wang, Y.~Zhou, and H.~Zhu, ``Client selection for federated learning with label noise,'' \emph{IEEE Transactions on Vehicular Technology}, pp. 2193--2197, 2021.

\bibitem{fang2022robust}
X.~Fang and M.~Ye, ``Robust federated learning with noisy and heterogeneous clients,'' in \emph{Proceedings of the IEEE/CVF Conference on Computer Vision and Pattern Recognition}, 2022.

\bibitem{xu2022fedcorr}
J.~Xu, Z.~Chen, T.~Q. Quek, and K.~F.~E. Chong, ``Fedcorr: Multi-stage federated learning for label noise correction,'' in \emph{Proceedings of the IEEE/CVF conference on computer vision and pattern recognition}, 2022.

\bibitem{yang2022robust}
S.~Yang, H.~Park, J.~Byun, and C.~Kim, ``Robust federated learning with noisy labels,'' \emph{IEEE Intelligent Systems}, pp. 35--43, 2022.

\bibitem{zhang2023noise}
J.~Zhang, D.~Lv, Q.~Dai, F.~Xin, and F.~Dong, ``Noise-aware local model training mechanism for federated learning,'' \emph{ACM Transactions on Intelligent Systems and Technology}, pp. 1--22, 2023.

\bibitem{zeng2022clc}
B.~Zeng, X.~Yang, Y.~Chen, H.~Yu, and Y.~Zhang, ``Clc: A consensus-based label correction approach in federated learning,'' \emph{ACM Transactions on Intelligent Systems and Technology (TIST)}, pp. 1--23, 2022.

\bibitem{tsouvalas2024labeling}
V.~Tsouvalas, A.~Saeed, T.~Ozcelebi, and N.~Meratnia, ``Labeling chaos to learning harmony: Federated learning with noisy labels,'' \emph{ACM Transactions on Intelligent Systems and Technology}, pp. 1--26, 2024.

\bibitem{wang2020principled}
T.~Wang, J.~Rausch, C.~Zhang, R.~Jia, and D.~Song, ``A principled approach to data valuation for federated learning,'' \emph{Federated Learning: Privacy and Incentive}, 2020.

\bibitem{li2023data}
W.~Li, S.~Fu, F.~Zhang, and Y.~Pang, ``Data valuation and detections in federated learning,'' \emph{arXiv preprint arXiv:2311.05304}, 2023.

\bibitem{bagdasaryan2020backdoor}
E.~Bagdasaryan, A.~Veit, Y.~Hua, D.~Estrin, and V.~Shmatikov, ``How to backdoor federated learning,'' in \emph{International conference on artificial intelligence and statistics}.\hskip 1em plus 0.5em minus 0.4em\relax PMLR, 2020.

\bibitem{wang2020attack}
H.~Wang, K.~Sreenivasan, S.~Rajput, H.~Vishwakarma, S.~Agarwal, J.-y. Sohn, K.~Lee, and D.~Papailiopoulos, ``Attack of the tails: Yes, you really can backdoor federated learning,'' \emph{Advances in Neural Information Processing Systems}, 2020.

\bibitem{singh2016comparative}
P.~Singh and R.~Shree, ``A comparative study to noise models and image restoration techniques,'' \emph{International Journal of Computer Applications}, 2016.

\bibitem{tian2020deep}
C.~Tian, L.~Fei, W.~Zheng, Y.~Xu, W.~Zuo, and C.-W. Lin, ``Deep learning on image denoising: An overview,'' \emph{Neural Networks}, 2020.

\bibitem{villar2021deep}
A.~Villar-Corrales, F.~Schirrmacher, and C.~Riess, ``Deep learning architectural designs for super-resolution of noisy images,'' in \emph{ICASSP 2021-2021 IEEE International Conference on Acoustics, Speech and Signal Processing (ICASSP)}.\hskip 1em plus 0.5em minus 0.4em\relax IEEE, 2021.

\bibitem{baradad2021learning}
M.~Baradad~Jurjo, J.~Wulff, T.~Wang, P.~Isola, and A.~Torralba, ``Learning to see by looking at noise,'' \emph{Advances in Neural Information Processing Systems}, 2021.

\bibitem{paul2021deep}
M.~Paul, S.~Ganguli, and G.~K. Dziugaite, ``Deep learning on a data diet: Finding important examples early in training,'' \emph{Advances in Neural Information Processing Systems}, 2021.

\bibitem{hendrycks2019robustness}
D.~Hendrycks and T.~Dietterich, ``Benchmarking neural network robustness to common corruptions and perturbations,'' \emph{Proceedings of the International Conference on Learning Representations}, 2019.

\bibitem{krizhevsky2009learning}
A.~Krizhevsky, G.~Hinton \emph{et~al.}, ``Learning multiple layers of features from tiny images,'' 2009.

\bibitem{medmnistv2}
J.~Yang, R.~Shi, D.~Wei, Z.~Liu, L.~Zhao, B.~Ke, H.~Pfister, and B.~Ni, ``Medmnist v2-a large-scale lightweight benchmark for 2d and 3d biomedical image classification,'' \emph{Scientific Data}, 2023.

\bibitem{xiao2017/online}
H.~Xiao, K.~Rasul, and R.~Vollgraf, ``Fashion-mnist: a novel image dataset for benchmarking machine learning algorithms,'' 2017.

\bibitem{helber2019eurosat}
P.~Helber, B.~Bischke, A.~Dengel, and D.~Borth, ``Eurosat: A novel dataset and deep learning benchmark for land use and land cover classification,'' \emph{IEEE Journal of Selected Topics in Applied Earth Observations and Remote Sensing}, 2019.

\bibitem{le2015tiny}
Y.~Le and X.~Yang, ``Tiny imagenet visual recognition challenge,'' \emph{CS 231N}, 2015.

\bibitem{wei2021learning}
J.~Wei, Z.~Zhu, H.~Cheng, T.~Liu, G.~Niu, and Y.~Liu, ``Learning with noisy labels revisited: A study using real-world human annotations,'' \emph{arXiv preprint arXiv:2110.12088}, 2021.

\bibitem{trockman2022patches}
A.~Trockman and J.~Z. Kolter, ``Patches are all you need?'' \emph{arXiv preprint arXiv:2201.09792}, 2022.

\end{thebibliography}
\bibliographystyle{IEEEtran}

\end{document}